\documentclass[10pt,twocolumn,letterpaper]{article}

\usepackage[pagenumbers]{iccv} % To force page numbers, e.g. for an arXiv version

\usepackage{graphicx}
\usepackage{subcaption}
\definecolor{iccvblue}{rgb}{0.21,0.49,0.74}
\usepackage[pagebackref,breaklinks,colorlinks,allcolors=iccvblue]{hyperref}
\newcommand{\red}[1]{{\color{red}#1}}

\usepackage{multirow}
\usepackage{graphicx}

\usepackage{float}
\usepackage{booktabs}
\usepackage{amsmath}
\usepackage{amssymb}
\usepackage{mathtools}
\usepackage{amsthm}
\usepackage{comment}
\newcommand{\xjqi}[1]{\textcolor[rgb]{1,0,0}{{[xjqi: #1]}}}

\usepackage{caption}
\usepackage{enumitem}
\usepackage{array}
\usepackage[table]{xcolor}

\newcommand{\bx}{\mathbf{x}}

\usepackage{environ}
\usepackage[utf8]{inputenc} % allow utf-8 input
\usepackage[T1]{fontenc}    % use 8-bit T1 fonts
\usepackage{url}            % simple URL typesetting
\usepackage{booktabs}       % professional-quality tables
\usepackage{amsfonts}       % blackboard math symbols
\usepackage{nicefrac}       % compact symbols for 1/2, etc.
\usepackage{microtype}      % microtypography
\usepackage{xcolor}         % colors

\RequirePackage{xspace}
\makeatletter
\DeclareRobustCommand\onedot{\futurelet\@let@token\@onedot}
\def\@onedot{\ifx\@let@token.\else.\null\fi\xspace}

\def\eg{\emph{e.g}\onedot} 

\def\ie{\emph{i.e}\onedot}

\makeatother

\usepackage[capitalize]{cleveref}
\crefname{section}{Sec.}{Secs.}
\Crefname{section}{Section}{Sections}
\Crefname{table}{Table}{Tables}
\crefname{table}{Tab.}{Tabs.}
\Crefname{appendix}{Appendix}{Appendices}
\crefname{appendix}{Appendix}{Appxs.}
\newcommand{\mbf}[1]{\mathbf{#1}}

\usepackage{multirow}
\usepackage{multicol}
\usepackage{booktabs} 
\newcommand{\PreserveBackslash}[1]{\let\temp=\\#1\let\\=\temp}
\usepackage{multirow}
\usepackage{amsmath}
\usepackage{graphicx}
\newcolumntype{?}{!{\vrule width 0.6pt}}
\usepackage{booktabs}       % professional-quality tables
\usepackage[utf8]{inputenc} % allow utf-8 input
\usepackage[T1]{fontenc}    % use 8-bit T1 fonts

\usepackage{url}            % simple URL typesetting
\usepackage{booktabs}       % professional-quality tables
\usepackage{amsfonts}       % blackboard math symbols
\usepackage{nicefrac}       % compact symbols for 1/2, etc.
\usepackage{microtype}      % microtypography
\usepackage{xcolor}         % colors
\usepackage{booktabs}       % professional-quality tables
\usepackage{tabularx}
\usepackage{multirow}
\usepackage{multicol}
\usepackage{makecell}

\def\paperID{831} % *** Enter the Paper ID here
\def\confName{ICCV}
\def\confYear{2025}

\title{Equipping Vision Foundation Model with Mixture of Experts for Out-of-Distribution Detection}

\author{
Shizhen Zhao\textsuperscript{} 
\;\; Jiahui Liu\textsuperscript{} 
\;\; Xin Wen\textsuperscript{} 
\;\; Haoru Tan\textsuperscript{}  
\;\; Xiaojuan Qi
\\
\; The University of Hong Kong  \\
}

\begin{document}
    
\maketitle

\begingroup
\renewcommand\thefootnote{\dag}
% \footnotetext{Corresponding author}
\endgroup

\begin{abstract}
% Pre-trained vision foundation models have transformed many computer vision tasks. 
% Despite their strong ability to learn discriminative and generalizable features crucial for out-of-distribution (OOD) detection, their impact on this task remains underexplored.
% %
% Motivated by this gap, we systematically investigate representative vision foundation models for OOD detection. Our findings reveal that a pre-trained DINOv2 model, even without fine-tuning on in-domain (ID) data, naturally provides a highly discriminative feature space for OOD detection, surpassing existing state-of-the-art methods without requiring complex designs or additional fine-tuning. 
% Beyond this, we explore how fine-tuning foundation models on in-domain (ID) data can enhance both ID classification and OOD detection. However, we observe that this process often leads to performance degradation due to the limited availability of ID data. To mitigate this, we propose the Mixture of Feature Experts (MoFE) module, which partitions features into subspaces, effectively capturing complex data distributions and refining decision boundaries even with limited ID data. 
% Further, we introduce a Dynamic-$\beta$ Mixup strategy, which samples interpolation weights from a dynamic beta distribution. This adapts to varying levels of learning difficulty across categories, improving feature learning for more challenging categories.
% Extensive experiments demonstrate the effectiveness of our approach, significantly outperforming baseline methods. 
% %
% The code will be made publicly available.
% %

Pre-trained vision foundation models have transformed many computer vision tasks. 
Despite their strong ability to learn discriminative and generalizable features crucial for out-of-distribution (OOD) detection, their impact on this task remains underexplored.
Motivated by this gap, we systematically investigate representative vision foundation models for OOD detection. Our findings reveal that a pre-trained DINOv2 model, even without fine-tuning on in-domain (ID) data, naturally provides a highly discriminative feature space for OOD detection, achieving performance comparable to existing state-of-the-art methods without requiring complex designs. 
Beyond this, we explore how fine-tuning foundation models on in-domain (ID) data can enhance OOD detection. 
However, we observe that the performance of vision foundation models remains unsatisfactory in scenarios with a large semantic space. 
This is due to the increased complexity of decision boundaries as the number of categories grows, which complicates the optimization process.
To mitigate this, we propose the Mixture of Feature Experts (MoFE) module, which partitions features into subspaces, effectively capturing complex data distributions and refining decision boundaries. 
Further, we introduce a Dynamic-$\beta$ Mixup strategy, which samples interpolation weights from a dynamic beta distribution. This adapts to varying levels of learning difficulty across categories, improving feature learning for more challenging categories.
Extensive experiments demonstrate the effectiveness of our approach, significantly outperforming baseline methods. 
The project will be available at \url{shizhen-zhao.github.io/OOD_MoFE/}. 

\end{abstract}

\section{Introduction}

\begin{figure}[t]
    \centering
    % \subfloat[CLIP - coarse-grained]{
    %     \includegraphics[width=0.3\textwidth]{figures/figure1_feature_space_v4/clip_vast.jpg}
    %     \label{fig:clip_vast}
    % }
    \subfloat[Generalized Model based Approach]{
        \includegraphics[width=0.45\textwidth]{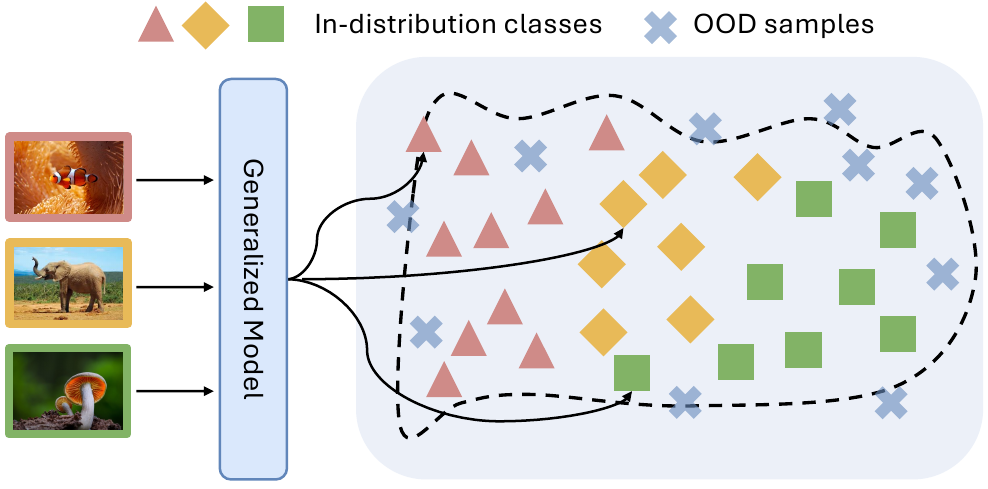}
        \label{fig:1a}
    }
    \hfill
    % \subfloat[CLIP - failure]{
    %     \includegraphics[width=0.3\textwidth]{figures/figure1_feature_space_v4/clip_Failure.jpg}
    %     \label{fig:clip_failure}
    % }
    % \vspace{-0.2cm}
    % \subfloat[DINOv2 - coarse-grained]{
    %     \includegraphics[width=0.3\textwidth]{figures/figure1_feature_space_v4/dinov2_vast.jpg}
    %     \label{fig:dinov2_vast}
    % }
    \subfloat[Expert-based Approach]{
        \includegraphics[width=0.45\textwidth]{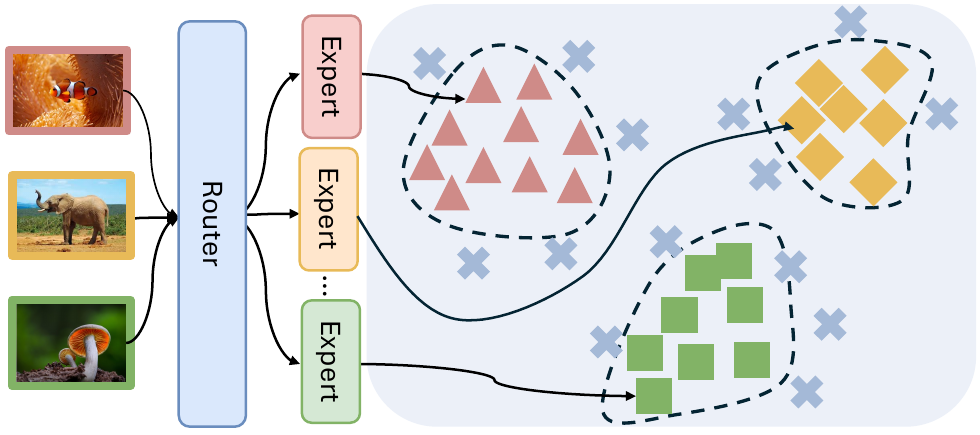}
        \label{fig:1b}
    }
    % \hfill
    % \subfloat[DINOv2 - failure]{\includegraphics[width=0.3\textwidth]{figures/figure1_feature_space_v4/dinov2_Failure.jpg}
    %     \label{fig:dinov2_failure}
    % }
\caption{{\bf Holistic comparison to previous philosophy.} (a) Traditional methods use a generalized model to project inputs onto a complex distribution; (b): Our approach leverages multiple experts to break the complex distribution into smaller ones, which leads to compact ID distribution and simplified decision boundary.} \label{fig:figure1}
% \vspace{-0.4cm}
\end{figure}

The task of out-of-distribution (OOD) detection~\citep{sehwag2021ssd,liang2018enhancing,hsu2020generalized,lee2018simple} aims to equip models with the capability to discern whether input images originate from unknown OOD classes or belong to in-domain (ID) classes. Mainstream OOD detection methods~\citep{Dream_OOD,tao2023nonparametric,du2022towards,lee2018training} focus on learning features and classifiers~\citep{sun2022dice,liang2018enhancing,sun2021react,wang2022vim} from ID data and then develop a score metric~\citep{hendrycks2019scaling,sun2022out,liu2020energy,hendrycks2016baseline} to determine whether a sample belongs to ID or OOD classes. Despite significant advancements, the fundamental challenge in OOD detection is establishing a feature space with high discriminative capacity that can effectively distinguish OOD samples from ID samples.
Recently, vision foundation models~\citep{oquab2023dinov2,radford2021clip,kirillov2023segany,singh2023effectiveness} trained on large-scale datasets have demonstrated the ability to learn robust and generalizable features, benefiting numerous tasks~\citep{depthanything,zhang2022glipv2,li2021grounded,tian2021vl}. This raises the question: \textit{with such powerful models and feature representations, does OOD detection remain a problem?}

Although several studies~\cite{Wang_2023_ICCV, Esmaeilpour_aaai_2022, ming2022delving, OOD_CLIP_IJCV_2024} have explored the use of foundation models for OOD detection, most focus on improving the performance of vision-language models like CLIP~\cite{radford2021clip}, while other foundation models, such as DINOv2, remain largely unexamined.
In this study, we systematically investigate the feature spaces of different representative pre-trained foundation models, including vision-language models (e.g., CLIP) and self-supervised models (e.g., DINOv2), in the context of OOD detection. Our results reveal that DINOv2 provides the most discriminative feature space, enabling effective OOD detection without any fine-tuning. Notably, using a simple KNN metric~\cite{sun2022out}, DINOv2 achieves performance comparable to more complex methods, establishing a strong baseline for further research.

While vision foundation models~\cite{oquab2023dinov2,radford2021clip,Wang_2023_ICCV,Nie_2024_ICLR} have achieved impressive performance in OOD detection, there is still room for improvement, particularly on in-domain data with large semantic spaces~\cite{tan2023datapruneinfomax,10262344,tan2024saco,tan2024ensembleqap,tan2023movingonesampleout,tan2023semanticdiffusion} (e.g., 29.27\% FPR95 on the ImageNet-1K OOD benchmark~\cite{sun2022out}). This prompts us to investigate whether foundation models can be further optimized by leveraging available ID data. 
However, as the number of semantic classes increases, the complexity of the decision boundaries required to distinguish between ID and OOD data grows as well~\cite{Huang_2021_CVPR,Li_2023_ICCV, fginference}. 
This heightened complexity creates challenges when fine-tuning foundation models on limited ID data.
Previous methods (\textit{e.g.} MOS~\cite{Huang_2021_CVPR}) decouple the complex space into simpler subspaces from the perspective of loss, which eases the optimization process and simplifies the decision boundaries. 
In this study, we tackle the problem orthogonally from the model perspective by designing a new Mixture of Experts (MoE) architecture to more thoroughly disentangle complex ID distribution.

To address this issue, rather than directly optimizing the whole feature space~\cite{liu2024can,liu2023mars3d,lyu2024total,chang2024matters} (\cref{fig:figure1}(a)),  we propose a Mixture-of-Feature-Expert (MoFE) module, which utilizes multiple experts, and each expert specializes in a specific subspace and optimizes it accordingly (\cref{fig:figure1}(b)).
MoFE operates by partitioning the original feature space into $K$ subspaces based on semantics and feature similarities within the ID dataset. 
Each subspace is assigned to a dedicated expert, and a router assigns samples to the appropriate expert based on these partitions. 
Different from previous studies~\cite{switchtransformer,moe}, we use the [CLS] token as the input to the router network, since it encapsulates the semantic feature of the whole image. 
During training, the router is supervised by the partition assignments to ensure accurate sample-to-expert mapping. 
Each expert focuses solely on optimizing features within its designated partition, which helps prevent interference between features from different partitions.
The results show that our approach significantly surpasses the previous approaches by a large margin (see \cref{tab:in1k_comp}), revealing the importance of learning expert models for OOD tasks.

Additionally, given that data augmentation has been shown to enhance generalization for OOD, we introduce a novel Mixup data augmentation strategy to further improve feature learning, which is better suited for advanced vision foundation models. 
Our design is based on the observation that different categories exhibit varying levels of discriminativeness with features from vision foundation models. 
In the original feature space, some categories show high discriminativeness, while others do not. 
For categories that are already well-represented, synthesizing dissimilar samples via vanilla Mixup~\cite{mixup} can blur the decision boundary between ID and OOD, leading to degraded performance. Thus, unlike existing Mixup strategies that treat all categories equally~\cite{wu2022palet,tan2021proxygraph,tan2025diffin}, our approach makes Mixup weight sampling category-dependent by adjusting the sampling distribution (\textit{i.e.} beta distribution) dynamically, taking into account their discriminativeness. 

Our major {\bf contributions} can be summarized as follows:
\begin{itemize}[leftmargin=*]
\item We design a novel MoFE module to tailor pre-trained vision foundation models for OOD detection. This approach reduces the difficulty of fitting complex data distributions from limited data and eases the optimization process.

\item We explore the effectiveness of the raw feature spaces from various vision foundation models for OOD detection. Through analysis, we leverage DINOv2 with simple scoring metric to establish a strong baseline. Additionally, we designed a Dynamic-$\beta$ Mixup that is better suited for advanced vision foundation models.

\item  Our extensive experimental results demonstrate the effectiveness of the proposed model, achieving significant improvements over several competitive baseline methods on standard benchmarks. 

\end{itemize}

\section{Related Work}

\paragraph{Out-of-Distribution Detection} The goal of OOD detection is to detect OOD images from the test dataset (containing both ID and OOD images). 
Designing the score function is the most popular method in OOD detection tasks. The scores are mainly derived from three sources: the probability~\citep{hendrycks2016baseline,hendrycks2019scaling}, the logits~\citep{hendrycks2019scaling,liu2020energy}, and the feature~\citep{lee2018simple,ndiour2020out}.
Some studies~\citep{Khalid,wang2022partial,sehwag2021ssd} focus on leveraging contrastive learning to enhance the feature representation.
Other studies show that synthesizing pseudo samples~\citep{du2022towards,sehwag2021ssd,tack2020csi,tack2020csi} as OOD instances is also a promising approach to make the feature space more compact. 
The methods~\cite{Li_2023_ICCV, fginference, Huang_2021_CVPR} are the most relevant to ours, which also break the semantic space into smaller ones. 
Different from these approaches, in our design, we propose a novel MoE module, with each expert exclusively concentrating on optimizing features within its specific partition.
Our results demonstrate that our approach outperforms them by a substantial margin.  

\paragraph{OOD Detection with Foundation Models} There are some existing OOD detection methods~\citep{Wang_2023_ICCV,Esmaeilpour_aaai_2022,ming2022delving,OOD_CLIP_IJCV_2024,Nie_2024_ICLR,miyai2023locoop} leveraging foundation models. 
Maximum Concept Matching (MCM)~\citep{ming2022delving} proposes a simple yet effective zero-shot OOD detection method by aligning visual features with textual concepts.
%
% ~\citep{OOD_CLIP_IJCV_2024} further investigates the effect of fine-tuning on OOD detection in large vision-language models and the MCM score is highlighted as effective.
%
Some other studies~\citep{Wang_2023_ICCV,Nie_2024_ICLR} explore negative prompts to learn the diversity of negative features,  enabling more accurate detection of OOD samples. 
Although these studies have made great progress by leveraging CLIP to enhance the performance in existing benchmarks, they only explore and fine-tune CLIP. 
In our studies, we explore different foundation models and explore a better fine-tuning paradigm. 

\paragraph{Mixture of Experts} Mixture of Experts has been studied independently in both computer vision~\citep{vmoe,mixermoe,limoe,wu2025mixtureofscores} and natural language processing~\citep{moe,gshard,switchtransformer,sparseupcycle}.
These works are studied in the context of conditional computation, which is to increase the number of model parameters without a proportional increase in computational cost. 
Currently, some studies~\citep{moe_data_conflict,moe_spe} explore improving expert specialization and leveraging MoE to mitigate data conflict problems, where some data might interfere with each other. 
In our study, we introduce MoFE to the out-of-distribution task in the context of foundation models and build specialized OOD detectors for different feature subspaces.

\section{Pilot Study}

In this section, we first introduce preliminaries for the OOD detection task in \cref{sec:prelim}. 
Then, we explore the impact of foundation models on OOD detection performance and analyze their strengths and weaknesses in \cref{sec:pre_analysis}. 

\subsection{Preliminaries}
\label{sec:prelim}

We consider supervised multi-class classification, where $\mathcal{X}$ represents the input image space and $\mathcal{Y}=\{1,2,...,C\}$ represents the label space. The training dataset $\mathbb{D}_{in} = \{(\bx_i, y_i)\}_{i=1}^n$ is drawn independently and identically distributed (\emph{i.i.d.}) from the joint data distribution $P_{\mathcal{X}\mathcal{Y}}$. Let $\mathcal{P}_\text{in}$ denote the marginal distribution on $\mathcal{X}$. Let $f: \mathcal{X} \mapsto \mathbb{R}^{|\mathcal{Y}|}$ be a neural network trained on samples drawn from $P_{\mathcal{X}\mathcal{Y}}$ to output a logit vector, which is used to predict the label of the input sample.

\noindent\textbf{Out-of-distribution Detection.} When deploying a machine learning model in real-world scenarios, it is crucial for a reliable classifier not only to accurately classify known in-distribution (ID) samples, but also to recognize any out-of-distribution (OOD) inputs as "unknown". This can be accomplished by incorporating an OOD detector alongside the classification model $f$. OOD can be formulated as a binary classification task. During testing, the objective is to determine whether a sample $\bx \in \mathcal{X}$ belongs to $\mathcal{P}_\text{in}$ (ID) or not (OOD). This decision can be made using a scoring metric $S(\bx)$:
\begin{align}
\label{eq:threshold}
	G_{\lambda}(\*x)=\begin{cases} 
      \text{ID} & S(\bx)\ge \lambda \\
      \text{OOD} & S(\bx) < \lambda 
   \end{cases},
\end{align}
where samples with higher scores $S(\bx)$ are classified as ID and vice versa, and  $\lambda$ is the threshold. 
Some typically used metrics $S(\bx)$ include MSP~\citep{hendrycks2016baseline}, MaxLogit~\citep{hendrycks2019scaling}, Energy~\citep{liu2020energy} and KNN~\citep{sun2022out}.
%In practice, OOD is often defined by a distribution that simulates unknowns encountered during deployment time, such as samples from an irrelevant distribution {whose label set has no intersection with $\mathcal{Y}$ and therefore should not be predicted by the model}.

% Recently, some methods have sought to address the issue of hard-to-distinguish OOD samples by leveraging generalizable representations learned by CLIP [10], an openworld language-vision model trained on datasets with enormous volumes, such as Laion-2B [36].

\begin{figure}[h]
\centering
\includegraphics[width=0.9\linewidth]{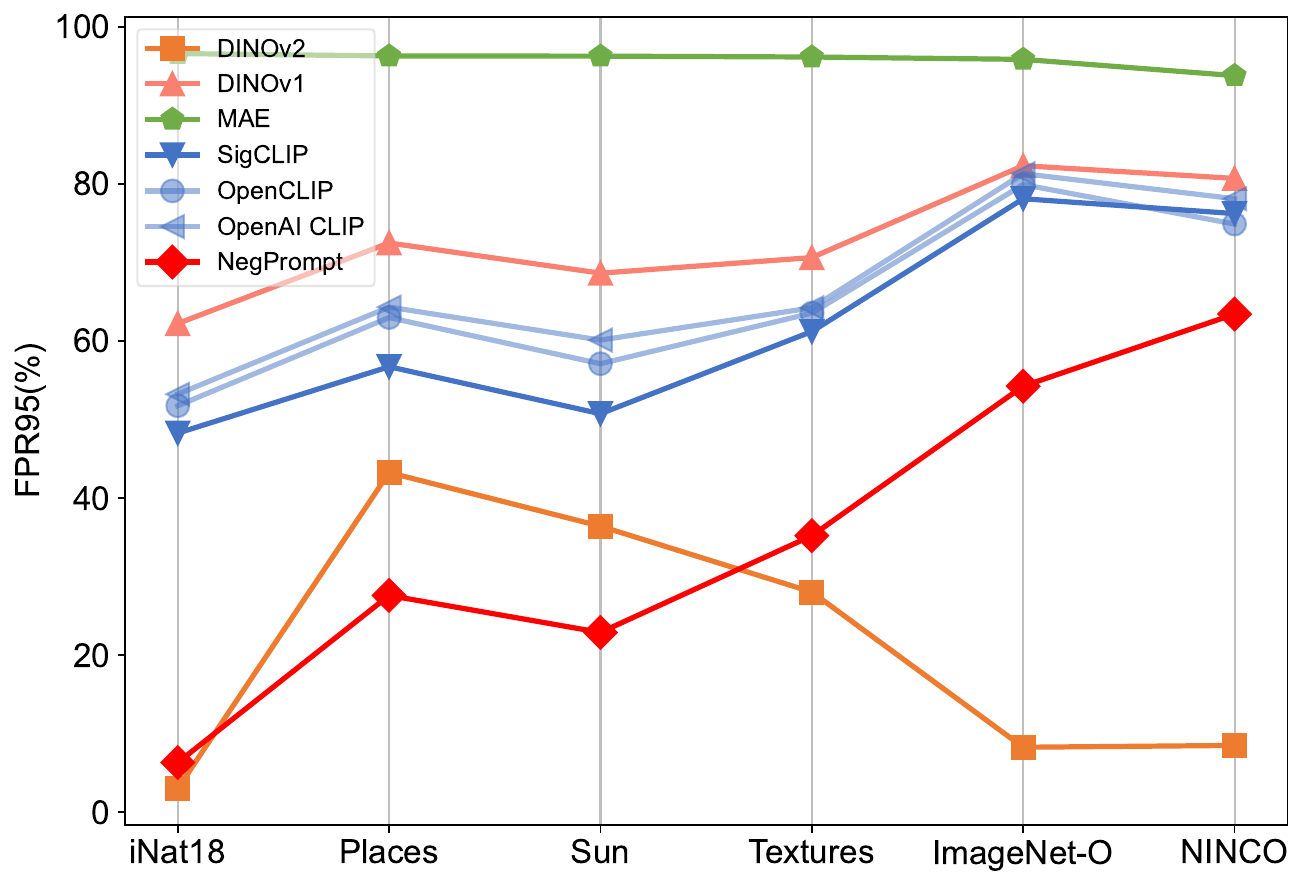}
\vspace{-0.1cm}
\caption{Performance of vision foundation models across different OOD splits. The evaluation metric is FPR95, with lower values indicating better performance.} \label{fig:Performance_foundation}
\vspace{-0.4cm}
\end{figure}

\subsection{Evaluation of Vision Foundation Models}
\label{sec:pre_analysis}

Although several studies~\cite{Wang_2023_ICCV, Esmaeilpour_aaai_2022, ming2022delving, OOD_CLIP_IJCV_2024} have explored the use of foundation models for OOD detection, they focus solely on vision-language foundation models such as CLIP~\cite{radford2021clip}. 
Beyond CLIP, the community offers a variety of vision foundation models that provide robust raw feature space.
This development has inspired us to re-examine which vision foundation model is best suited for OOD detection. 
In this section, we aim to investigate and analyze various pre-trained vision foundation models as effective OOD detectors without fine-tuning.

\begin{figure}[t]
    \centering
    % \subfloat[CLIP - coarse-grained]{
    %     \includegraphics[width=0.3\textwidth]{figures/figure1_feature_space_v4/clip_vast.jpg}
    %     \label{fig:clip_vast}
    % }
    % \hfill
    \subfloat[CLIP]{
        \includegraphics[width=0.23\textwidth]{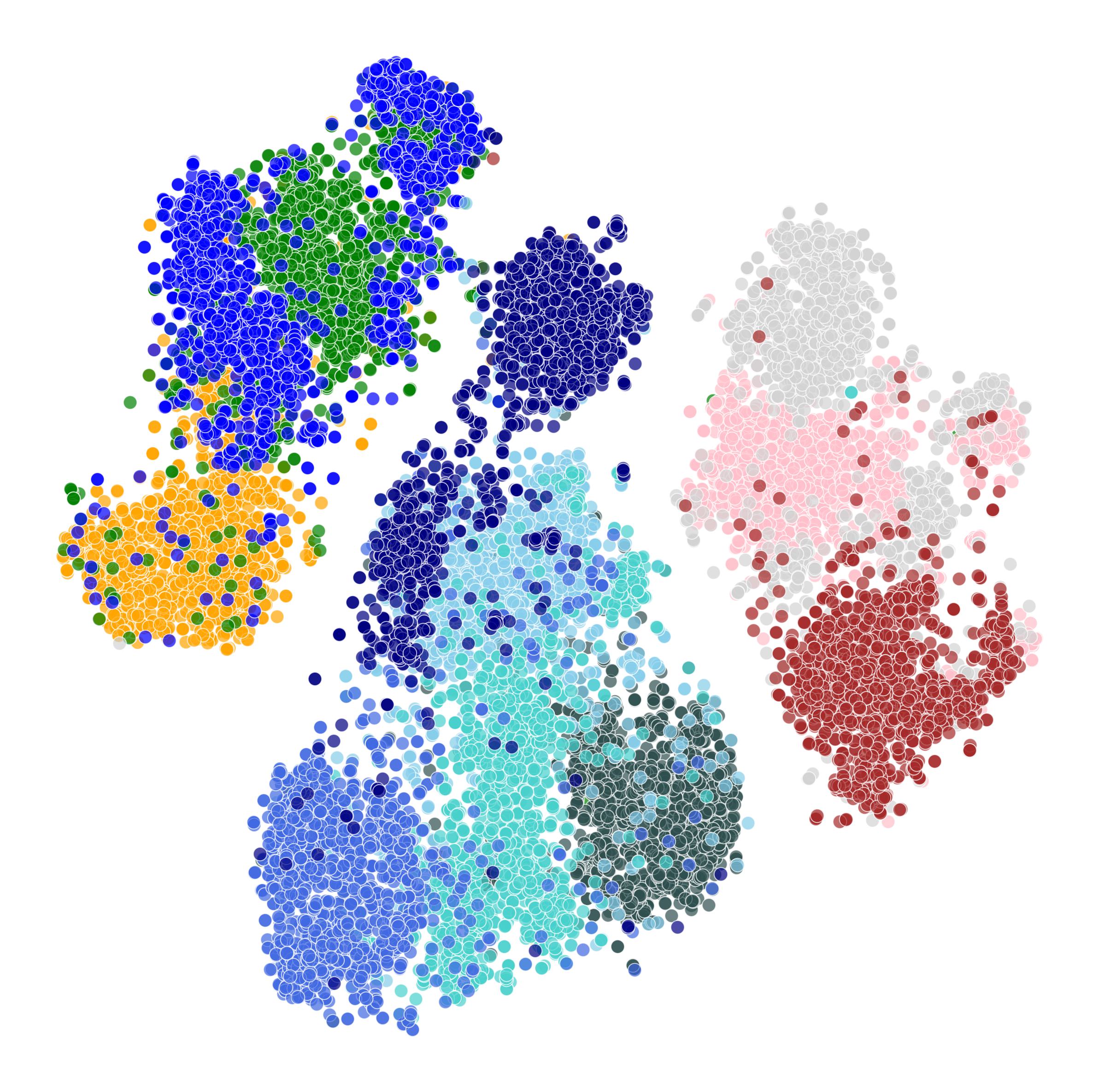}
        \label{fig:clip_finegrained}
    }
    % \hfill
    % \subfloat[CLIP - failure]{
    %     \includegraphics[width=0.3\textwidth]{figures/figure1_feature_space_v4/clip_Failure.jpg}
    %     \label{fig:clip_failure}
    % }
    % \vspace{-0.2cm}
    % \subfloat[DINOv2 - coarse-grained]{
    %     \includegraphics[width=0.3\textwidth]{figures/figure1_feature_space_v4/dinov2_vast.jpg}
    %     \label{fig:dinov2_vast}
    % }
    % \hfill
    \subfloat[DINOv2]{
        \includegraphics[width=0.23\textwidth]{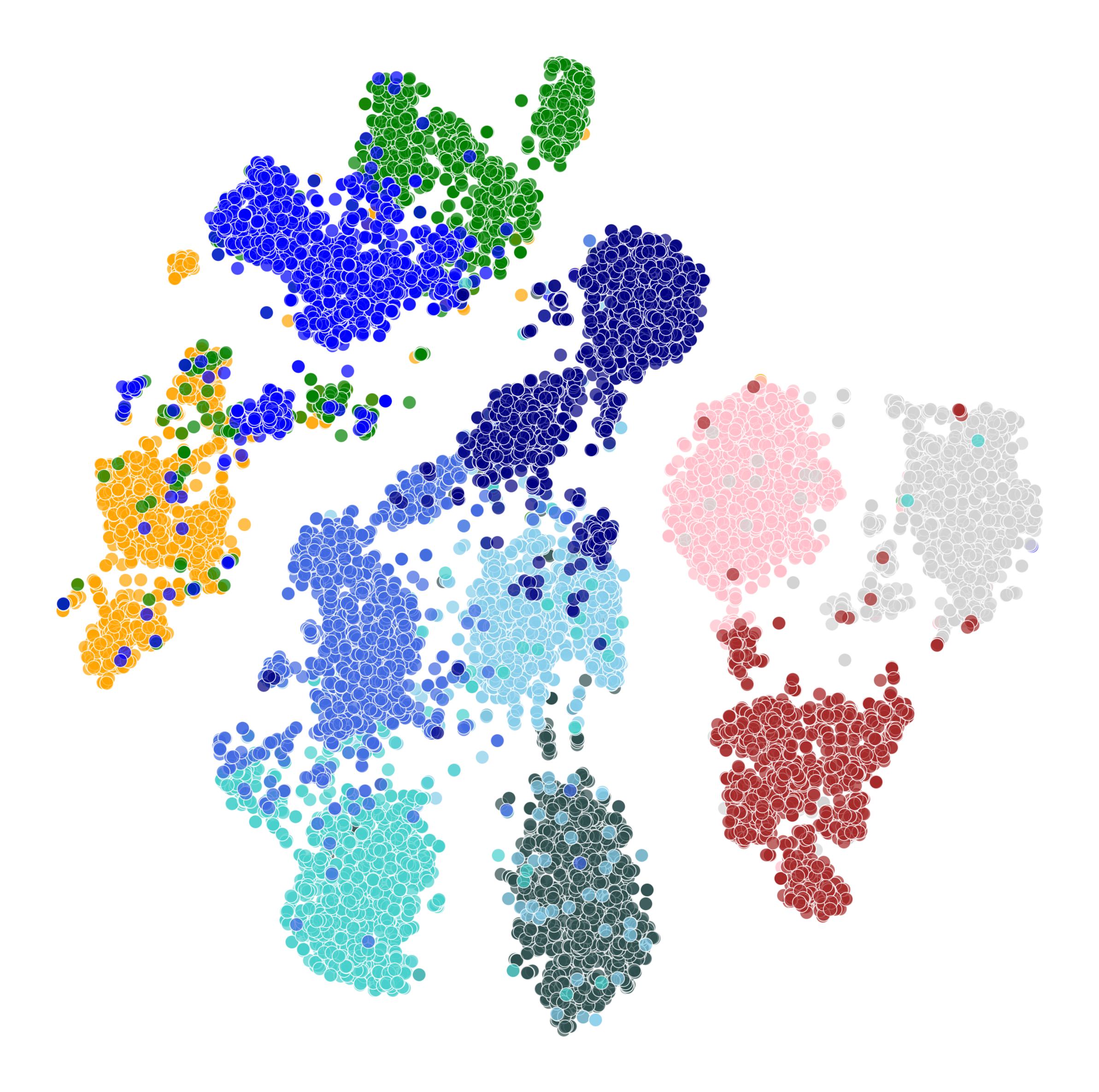}
        \label{fig:dinov2_finegrained}
    }
    % \hfill
    % \subfloat[DINOv2 - failure]{\includegraphics[width=0.3\textwidth]{figures/figure1_feature_space_v4/dinov2_Failure.jpg}
    %     \label{fig:dinov2_failure}
    % }
    \caption{{\bf Feature Visualization for Foundation Models.} For fine-grained feature visualization, we randomly select fine-grained categories under 3 different super classes from ImageNet-1k.}
    \label{fig:feature_space}
% \vspace{-0.4cm}
\end{figure}

% \vspace{-0.2cm}
\noindent \textbf{Experimental Setup.} 
We perform our evaluation on a challenging OOD detection benchmark that utilizes ImageNet-1K as ID data and selects samples from iNaturalist18, Sun, Places, and Textures as OOD samples. 
We also include two challenging OOD test sets: ImageNet-O~\cite{hendrycks2021nae} and NINCO~\cite{bitterwolf2023ninco}. 
We choose several representative vision foundation models, namely DINOv1~\cite{caron2021emerging}, DINOv2~\citep{oquab2023dinov2}, MAE~\cite{MaskedAutoencoders2022}, SigCLIP~\cite{zhai2023sigmoid}, OpenCLIP~\cite{ilharco_gabriel_2021_5143773}, and OpenAI CLIP~\cite{radford2021clip}.
For fair comparison, we use the ViT-B as the architecture of these models.  
The scoring functions for DINOv1, DINOv2, and MAE are set to KNN~\cite{sun2022out}. 
For the CLIP series, we report the best results among four scoring functions (MSP, MaxLogit, Energy and KNN). 
Without any model tuning, we directly use the features extracted from these models for OOD detection evaluation to assess whether they are already sufficiently capable of OOD detection. 
To emphasize the significance of our findings, we also compare them with the state-of-the-art method (NegPrompt~\cite{li2024learning}) that involve fine-tuning an ImageNet pre-trained model on the ID dataset. 
Additionally, we conduct further verification using datasets beyond ImageNet-1k as ID data in \cref{sec:more_pilot} and the conclusion align with the experiments using ImageNet-1k.

\noindent \textbf{Result Analysis.} 
\noindent\textbf{(1)} With traditional score metrics (\emph{i.e} KNN), DINOv2 can outperform all other foundation models by a large margin.
DINOv2+KNN shows the best results where the average FPR95 is 29.27\% in the first four test sets, 8.9\% in the latter challenging test sets, while SigClIP only achieves 54.23\% and 77.17\%. 
This is potentially because DINOv2 leverages advanced self-supervised learning: iBot~\citep{zhou2021ibot}, which is a Mask Image Modeling (MIM) pretask for facilitating models to capture image details, and contrastive learning objective~\citep{caron2021emerging} that enhances the feature discriminativeness. 
\noindent\textbf{(2)} Without any fine-tuning, DINOv2 achieves performance comparable to the more complex method (\textit{i.e.} NegPrompt) in the first four test sets. 
Notably, DINOv2+KNN still significantly outperforms NegPrompt on challenging test sets by 49.94\%. 
The reason is that these two datasets contain images that are extremely similar to the ID categories.
However, the paradigm of CLIP only provides image-level textual supervision without a supervision signal to retain detailed image information. 
Therefore, CLIP always fails in some fine-grained tasks, while DINOv2 consistently performs much better.
As shown in \cref{fig:clip_finegrained} and \cref{fig:dinov2_finegrained}, where we randomly select 11 fine-grain categories under 3 different super classes, DINOv2 provide more discriminative boundaries, while CLIP can not.  

%
% Due to the limited space, we include the results in Tab.~\red{7} in Appendix.
% %
% We compare the OOD detection performance of foundation models without fine-tuning and SOTA ImageNet pre-trained models with fine-tuning, where we can observe that:

\begin{figure*}[t]
\centering
\includegraphics[width=\linewidth]{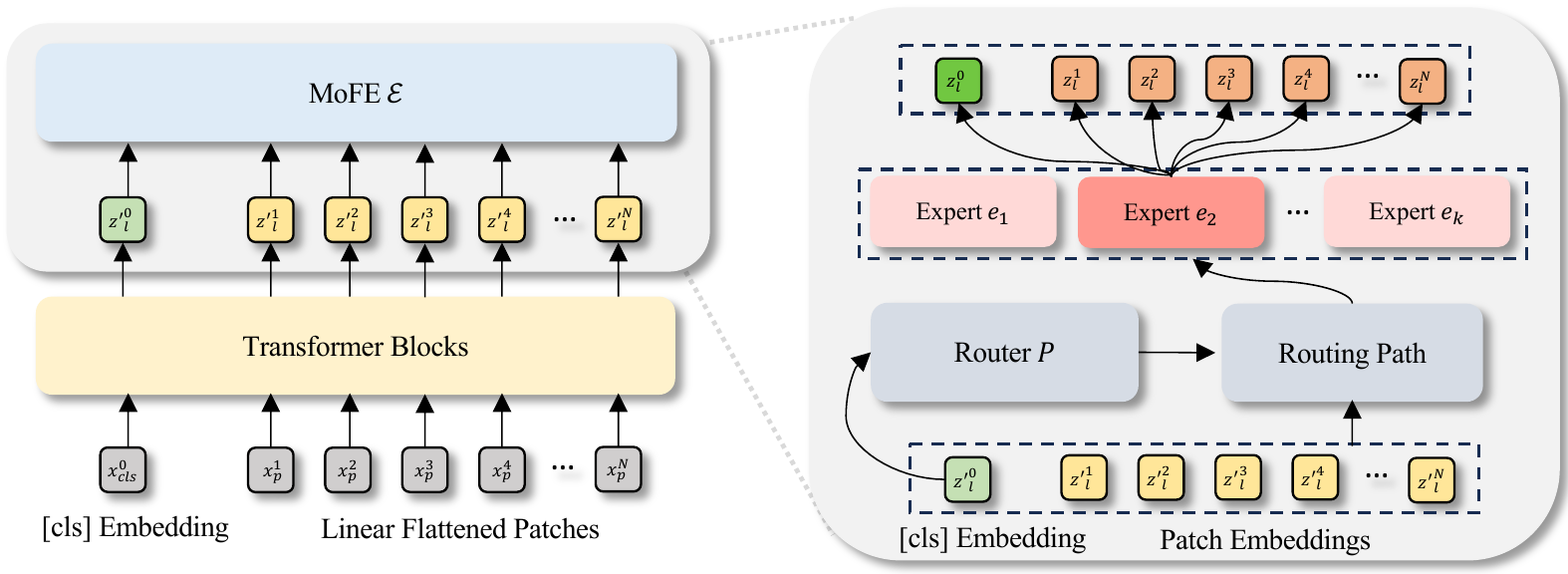}
\caption{\textbf{Illustration of our proposed Mixture of Feature Experts (MoFE).} 
MoFE decomposes the large semantic space into multiple subspaces and each expert specializes in a specific subspace.
Specifically, the image patches and the class token are input to obtain the preliminary patch embeddings and class embedding. 
A router is employed to determine the expert to further process the embeddings, and the input of the router is the class embedding.
Finally, we apply associated experts to refine the class embeddings and the patch embeddings.
We use the class embeddings output by MoFE and conduct the OOD detection in the corresponding subspace. 
}
\vspace{-0.4cm}
\label{fig:framework}
\end{figure*}

\noindent \textbf{Further Challenges in OOD using Foundation models.} 
In summary, DINOv2, without requiring any fine-tuning, can already function as a high-performing OOD detector, surpassing previous approaches and underscoring \textit{the importance of discriminative and generalizable features for OOD detection}. 
However, foundation models still have room to improve and cannot generalize well across the entire feature space.  
\textbf{(1)} Though there is a consensus that fine-tuning on the ID data can improve OOD performance~\citep{vaze2022openset,chen_2020_ECCV,hendrycks2019scaling,tack2020csi}, we find that this doesn't hold in the context of foundation models, particularly on in-domain data with large semantic spaces. For instance, when we fine-tune DINOv2 on ImageNet-1K ID data and evaluate the fine-tuned model, the performance declines on three out of the four OOD datasets. The implementation details of this finetuning can be referred to \cref{sec:Naive_finetuning}.
\textbf{(2)} Besides, as shown in \cref{fig:clip_failure} and \cref{fig:dinov2_failure} in Appendix, we also show some failure examples, where the models exhibit particularly poor feature discriminability, hindering effective OOD detection.  
%
% These findings motivate us to design specialized fine-tuning methods for vision foundation models to achieve better OOD performance.

% \vspace{-0.2cm}

\section{Method}

% \xjqi{add transition here. We further study whether fine-tuning can  improve performance. say sth about directly fine-tuning poor performance. Then, motivate your approach 1. mixture of feature experts and 2. data augmentation}

This section introduce our proposed methods for finetuning vision foundation models to enhance the OOD detection ability, which includes a Mixture of Feature Expert module in \cref{sec:mode} and a Dynamic-$\beta$ Mixup data augmentation strategy in \cref{sec:mixup}. 

\subsection{Mixture of Feature Experts}
\label{sec:mode}

% \xjqi{add a summary?? with some high-level ideas about what this module can help}

As shown in~\cref{fig:framework}, we propose Mixture of Feature Experts (MoFE), which divides the complex semantic space into multiple subspaces and each expert specializes in a specific subspace. 
Each expert can tackle an easier problem instead of conducting OOD detection on a complicated distrubution, which eases the optimization process while maintaining the generalizability of features. Below presents the detailed configuration of MoFE. 
%
% Besides, we explain the difference with the traditional MoE in Appendix \red{A.4}.

Given an RGB image $\mathbf{v} \in \mathbb{R}^{H \times W \times 3}$, where $H$ and $W$ are the origin resolution,  we reshape the image $\mbf{x} \in \mathbb{R}^{H \times W \times C}$ into a sequence of flattened 2D patches $\mbf{x}_p \in \mathbb{R}^{N \times (P^2 \cdot C)}$,  $C$ is the number of channels, $(P, P)$ is the resolution of each image patch.
Next, we flatten the patches and map to $D$ dimensions with a trainable linear projection $\textbf{E} \in \mathbb{R}^{(P^2 \cdot C) \times D}$.
A learnable embedding is prepended to the sequence of embedded patches ($\mbf{z}_0^0=\mbf{x}_\text{cls}^0$) and position embeddings are added to the patch embeddings ${E}_{pos}  \in \mathbb{R}^{(N + 1) \times D} \label{eq:transformer}$.
Then we input these embeddings to multiple transformer blocks.
The output is processed by a MoFE layer to obtain the domain-specific features. 
This process is expressed as:
\begin{align}
    \mbf{z}_0 &= [ \mbf{x}_\text{cls}^0; \, \mbf{x}^1_p \mbf{E}; \, \mbf{x}^2_p \mbf{E}; \cdots; \, \mbf{x}^{N}_p \mbf{E} ] + \mbf{E}_{pos},  \label{eq:transformer}  \\
    \mbf{z^\prime}_\ell &= \text{Transformer}(\mbf{z}_{\ell-1}) + \mbf{z}_{\ell-1},  \ell=1\ldots L, \\
    \mbf{z}_{\ell} &=\mathrm{MoFE}(\mathrm{LN}(\mbf{z^{\prime}}_{\ell}))+\mbf{z^{\prime}}_{\ell},  \label{eq:moe}\\
    \mbf{F}&=\mathrm{LN}(\mathbf{z}_l). \label{eq:final_ln}
\end{align}
where $LN$ denotes the layer norm.

\noindent\textbf{MoFE Architecture.}
%，
% \textbf{MoE Forward \xjqi{Aritecture??}.} \xjqi{should this be MoE or MoFE, change it to MoFE would be better. Add the discuss relationships and differences of MoFE and MoE}  \xjqi{no need to mention MoE. you can directly describe the structure of MoFE. As shown in Figure x, MoFE contains .....}
%
The MoFE layer consists of multiple expert networks, each of which is a transformer block. 
As an initialization step, we replicate the transformer blocks from the final layer of a foundation model to form an ensemble of experts {\small $\mathcal{E}=[e_{1},e_{2},\cdots,e_{E}]$}. 
The router~\cite{moe} is a linear layer that predicts the probability of each token being assigned to each expert. 
Routing accuracy is crucial for MoFE. The key question is what should be used to determine the results of feature routing? We explore various approaches, such as reinitializing a routing token, averaging patch embeddings, or utilizing class embeddings. We ultimately find that using the class embedding achieves the best results. Although it is not the embedding output from the last layer of the network, it is sufficiently discriminative.
Therefore, we utilize the class embedding {\small $\mbf{z^\prime}_{l}^0$} as the input of the router. 
%
% Therefore, we utilize The input of the router is the class embedding {\small $\mbf{z^\prime}_{l}^0$} output by \cref{eq:transformer}. 
%
The router is a linear layer that predicts the probability of each token being assigned to each expert. We formulate as:
\begin{equation}
    \mathcal{P}(\mbf{z^\prime}_{l}^0)_i=\frac{e^{f(\mbf{z^\prime}_{l}^0)_i}}{\sum_j^E e^{f(\mbf{z^\prime}_{l}^0)_j}},
\end{equation}
\noindent where the router produces weight logits $f(\mbf{z^\prime}_{l}^0) = \mathbf{W}\cdot\mbf{z^\prime}_{l}^0$, which are normalized by the softmax function. $\mathbf{W}\in\mathbb{R}^{D\times E}$ represents the lightweight training parameters and $E$ represents the number of experts. 
After determining the experts by using the class embedding, we input all embedding including the patch embeddings and class embedding to the activated experts. 
Each embedding is processed by the top-$k$ experts with the highest probabilities, and the weighted sum is calculated based on
the softmax results of the probabilities:
\begin{equation}
    \mathrm{MoFE}(\mbf{z^\prime}_\ell)=\sum_{i=1}^k\mathcal{P}(\mbf{z^\prime}_{l}^0)_i\cdot\mathcal{E}(\mbf{z^\prime}_\ell)_i,
\end{equation}
where $\mathcal{E}$ represents the network of an expert~\cite{switchtransformer}. In our MoFE architecture, we route to only a single expert, thus $k=1$. 
We find that the router computation is reduced as we are only routing a token to a single expert and the performance does not increase when using more experts. 

\noindent\textbf{Feature Space Separation.} In MoFE, we aim to have different experts specialize in different subspaces.
Therefore, we propose to first separate the whole feature space into multiple subspaces so that each expert specializes in learning features within its subspace.
We use WordNet~\cite{wordnet}, which provides a good summary of the higher-level semantics of categories, to offer an initial partition of the subspace.
Since the semantic and visual similarities are not completely equivalent, we further refine the clustering using K-Means to adjust for the discrepancies.
Specifically, we extract feature representations $\mbf{z^\prime}_{l}^0$ for each training image. 
Then, we calculate the class prototypes by averaging the features of the images from each category. 
Finally, we perform a K-Means clustering on categorical feature prototypes.
The initial cluster centers are determined by the centroids of clusters, which are divided according to the original semantic space.
After determining the clustering results, we assign different experts to different clusters, with samples from each category being routed to the corresponding expert model.

% Therefore, we explicitly define the route path for each sample.
%
% We determine the initial clustering centers based on WordNet semantic information. Each class is associated with a synset in WordNet, from which we can build the taxonomy as a hierarchical tree. We average the class features of all sub-categories from these super categories. Therefore, the class clustering is consistent across multiple runs.  
%

\noindent\textbf{MoFE Training.} 
%
% We separate the last $M$ transformer blocks as the MoFE layer, where $M=1$ by default. 
We replace the final transformer block with a MoFE layer. Each transformer block within MoFE is initialized by the original final transformer block.
We could set multiple layers as MoFE layers, but we find that using just the final layer achieves sufficiently good results.
Then we randomly initialize a router layer and use the class token as the input. 
We use the labels generated by the above clustering to supervise the routing:
%\vspace{-0.2cm}
\begin{equation}
    \mathcal{L}_\text{route} = - \frac{1}{N}\sum_{n=1}^N \sum_{i=1}^Ey^i\log (\mathcal{P}^i (\mbf{z^\prime}_{l}^0)).
%\vspace{-0.1cm}
\end{equation}
For each expert, we leverage the categories within the corresponding cluster as the positive samples, and the categories beyond the cluster as the negative ones. 
Assuming that the category cluster of the i$th$ expert contains $Q_i$ classes, we set the categories beyond the cluster as the $Q_i+1$ categories. The loss is designed as follows:
\begin{equation}
    \mathcal{L}_\text{expert} = - \frac{1}{N}\sum_{n=1}^N \sum_{i=1}^E \sum_{q=1}^{Q_i+1} y^i y^q \log (p_i^q (\mathbf{x})).
%\vspace{-0.1cm}
\end{equation}
In order to achieve the sample balance for each cluster, we control the ratio of positive and negative samples as 1:1 during training. Therefore, the overall loss of MoFE is:
\begin{equation}
    \mathcal{L}_\text{MoFE} = \mathcal{L}_\text{expert} + \mathcal{L}_\text{route}.
%\vspace{-0.1cm}
\end{equation}

\paragraph{Discussion.} 
MoFE is designed to address OOD issues under large-scale complex distributions. 
Similar to MOS~\citep{Huang_2021_CVPR}, we decouple complex distributions into simpler subspace.
However, while MOS approaches this solely from the perspective of the loss function, we approach it from the model perspective by assigning an expert model to each subspace. 
This allows each expert to focus on learning its assigned subspace, preventing interference between features from different partitions. 
To accurately assign expert models to different samples during inference, we have devised a new routing method that uses the [CLS] token as the input to the router network, since it encapsulates the semantic features of the entire image.
With these designs, MoFE outperforms MOS by a substantial margin (see \cref{tab:in1k_comp}) and the feature visualization shows that MoFE can achieve more compact ID distribution and clearer decision boundaries between ID and OOD samples (Appendix \cref{tab:motivation_mofe}).

\subsection{Dynamic-$\beta$ Mixup}
\label{sec:mixup}

% \xjqi{change the words following the introduction }
Data augmentation (\eg, Mixup~\citep{mixup,mixup_1}) has been proven to improve generalization during finetuning. 
Traditional Mixup~\citep{mixup,mixup_1} augment samples and transform labels by:
\begin{align}
\tilde{x} = \lambda x_i + (1-\lambda) x_j, \quad \tilde{y} = \lambda y_i + (1-\lambda) y_j,
\end{align}
where $\lambda \sim \text{Beta}(\sigma, \sigma)$.
%
% Since the probability density function of $\text{Beta}(\sigma, \sigma)$ is symmetric about 0.5, for simplicity of explanation in our implementation, when the beta sampling exceeds 0.5, lambda is set to 1-beta.
%
$\lambda$ is the interpolation weight for generating new augmented samples. 
We observe that different categories exhibit varying levels of discriminativeness initialized by vision foundation models, as shown in Appendix \cref{fig:clip_failure} and \cref{fig:dinov2_failure}.
For categories that are already well-represented, synthesizing dissimilar samples via vanilla Mixup can blur the decision boundary between ID and OOD, leading to degraded performance (Fig.~\red{4} in Appendix). 
Therefore, we dynamically adjust the $\text{Beta}$ distribution according to the feature discriminativeness per category. 
The reason is that when features of $x_i$ are discriminative enough, a small $\lambda$, which leads to a dissimilar sample, is not necessary for their representation learning. 
Instead, we should leverage similar samples from a large $\lambda$ for building smooth decision boundaries. 
On the contrary, when features of a category show poor discriminativeness, we should set a relatively small $\lambda$ to ease the feature learning. 
We use the accuracy of the validation set to measure the discriminativeness.
Therefore, we set $\lambda$ as:
\begin{equation}
    \lambda \sim \text{Beta}(\sigma, \sigma) \ \text{for} \ \sigma=1-w*s,
\end{equation}
% \begin{align}
% \label{eq:threshold}
% 	\lambda=\begin{cases} 
%       \hat{\lambda} & \hat{\lambda} < 0.5 \\
%       1 - \hat{\lambda} & \hat{\lambda} \ge 0.5 
%    \end{cases},\hat{\lambda} \sim \text{Beta}(\sigma, \sigma) \ \text{for} \ \sigma=1-w*s
% \end{align}
where $w$ is a scaling factor and s denotes the corresponding category's accuracy on the validation set. 
Because the probability density function of $\text{Beta}(\sigma, \sigma)$ is symmetric about 0.5 and ranges from 0 to 1, we need to ensure that with a larger $s$, the probability of sampling larger values is greater. Therefore, we transform $\lambda$ as:
\begin{align}
\label{eq:threshold}
	\hat{\lambda}=\begin{cases} 
      \lambda & \lambda \ge 0.5 \\
      1 - \lambda & \lambda < 0.5 
   \end{cases}.
\end{align}
We determine the category difficulty at the beginning of the training and then update it during the training process.
In our implementation, $x_i$ is the training sample, and $x_j$ is the instance used to corrupt $x_i$.  
Therefore, we select the s from categories of $x_i$, and we select samples from different classes. 
Additionally, we empirically find that using vanilla Mixup~\citep{mixup,mixup_1} can cause feature norms to grow during finetuning vision foundation models (\ie, DINOv2), leading to performance degradation on the OOD task.
In order to restrain the growth of feature norms, we propose to add a regularization term to suppress the increase in feature norm:
\begin{equation}
    \mathcal{L}_\text{Mixup} = - \frac{1}{N}\sum_{n=1}^N \sum_{c=1}^C y^c\log (p^c(x)) + Reg(F^0),
%\vspace{-0.1cm}
\end{equation}
where {\small $C$} is the total number of categories, $Reg$ denotes a regularization method, {\small $F^0$} is the final class embeddings output by MoFE.  
By default, the regularization method has multiple choices, which can be $L_2$ norm or label smoothing. 
The final optimization objective is:
\begin{equation}
    \mathcal{L}_\text{final} = \mathcal{L}_\text{MoFE}  + \mathcal{L}_\text{Mixup}.
%\vspace{-0.1cm}
\end{equation}

\begin{table*}[t]
\centering
\renewcommand\arraystretch{1.2}
\resizebox{0.99\textwidth}{!}{
\begin{tabular}{llccccccccccc}
    \toprule
              \multicolumn{2}{c}{\multirow{2.5}{*}{\textbf{Method}}} & \multicolumn{2}{c}{\textbf{iNaturalist18}} & \multicolumn{2}{c}{\textbf{Places}} & \multicolumn{2}{c}{\textbf{Sun}} & \multicolumn{2}{c}{\textbf{Textures}}  & \multicolumn{2}{c}{\bf Average} & \multirow{2.5}*{\bf ID ACC$\uparrow$}  \\
    \cmidrule(r){3-4} \cmidrule(r){5-6} \cmidrule(r){7-8} \cmidrule(r){9-10} \cmidrule(r){11-12}        
    \multicolumn{2}{c}{}& \bf FPR95$\downarrow$ &     \bf AUROC$\uparrow$ & \bf FPR95$\downarrow$ & \bf AUROC$\uparrow$ & \bf FPR95$\downarrow$ & \bf AUROC$\uparrow$ & \bf FPR95$\downarrow$ & \bf AUROC$\uparrow$ & \bf FPR95$\downarrow$ & \bf AUROC$\uparrow$   \\
    \midrule    
    \multirow{7}[1]{*}{\rotatebox{90}{CLIP-Based}} 
    & \makecell[l]{Energy  \citep{liu2020energy}} &  65.00	&  87.17	& 57.40 & 87.32	& 46.43  & 91.17	& 57.40	 & 87.32	&  56.55	& 88.24	 & 79.39 \\
    & \makecell[l]{MSP  \citep{hendrycks2016baseline}} &40.89&	88.63&	65.81&	81.24&	67.90&	80.14&	64.96&	78.16&	59.89&	82.04 & 79.39 \\
    & \makecell[l]{MaxLogit  \citep{hendrycks2019scaling}} & 60.86	& 88.03	& 55.5 & 87.44	&  44.81& 91.16	& 52.25	 & 86.04	&  53.35	& 88.16	 &  \bf 79.39\\
    & \makecell[l]{MCM  \citep{ming2022delving}} 
    & 30.91	& 94.61	& 37.59& 	92.57& 	44.69& 	89.77& 	57.77& 	86.11& 	42.74& 	90.77 & 67.01 \\
    & \makecell[l]{CLIPN  \citep{Wang_2023_ICCV}} 
    & 23.94	& 95.27	& 26.17 & 	93.93 &  33.45 & 	92.28 & 	40.83 & 	90.93& 	31.10& 	93.10 & 68.53 \\
    & \makecell[l]{LSN  \citep{Nie_2024_ICLR}} & 21.56 & 95.83  & 34.48 & 91.25 & 26.32 &  94.35  &  38.54 & 90.42 & 30.22 & 92.96 & 71.89\\
    & \makecell[l]{NegPrompt  \citep{li2024learning}} & 6.32 & \bf 98.43  & 27.60 & 93.34 & 22.89 & \bf 95.55  &  35.21 & 91.60 & 23.01 & 94.81 & 66.84\\
    % & \cellcolor{gray!10}\emph{Ours} & \cellcolor{gray!10}\bf 17.19 & \cellcolor{gray!10}\bf 97.01  &  \cellcolor{gray!10}\bf 24.27 & \cellcolor{gray!10}\bf 94.35 & \cellcolor{gray!10}\bf 22.47  & \cellcolor{gray!10}94.27 & \cellcolor{gray!10}\bf 35.79 & \cellcolor{gray!10}\bf 91.45 & \cellcolor{gray!10}\bf 24.92 & \cellcolor{gray!10}\bf 94.27 & \cellcolor{gray!10}\bf 73.43 \\
        & \cellcolor{gray!10}\emph{Ours} & \cellcolor{gray!10}\bf 5.19 & \cellcolor{gray!10} 97.28  &  \cellcolor{gray!10}\bf 21.32 & \cellcolor{gray!10}\bf 94.69 & \cellcolor{gray!10}\bf 22.10  & \cellcolor{gray!10}95.17 & \cellcolor{gray!10}\bf 31.47 & \cellcolor{gray!10}\bf 92.15 & \cellcolor{gray!10}\bf 20.02 & \cellcolor{gray!10}\bf 94.89 & \cellcolor{gray!10} 68.56 \\

    \midrule 
    \multirow{5}[1]{*}{\rotatebox{90}{Dinov2-Based}}
    & \makecell[l]{MSP  \citep{hendrycks2016baseline}} & 25.02 	& 94.76	& 57.09 & 83.45	& 53.65  & 85.22	& 48.79 & 85.81	& 48.13	&  87.31	 & 86.01\\
    & \makecell[l]{MaxLogit  \citep{hendrycks2019scaling}} &  22.96	& 	94.59 & 59.21 & 78.41	& 54.52 & 81.80	& 48.17	 &84.16 	& 46.21	& 	84.74 & 86.01 \\
    & \makecell[l]{Energy  \citep{liu2020energy}} & 28.48	& 93.19 	&  65.88 & 74.49	& 61.54 & 78.71	&	53.29 & 	81.92 &  52.29	&  	82.07 &  86.01\\
    &\makecell[l]{KNN  \citep{sun2022out}} & 5.67 & 97.65  & 43.25 & 88.21 & 36.42 & 90.21  &  28.04 & 92.66 & 28.34 & 92.18 & 86.01\\
        &\makecell[l]{MOS~\citep{Huang_2021_CVPR}} & 5.01 &  97.85 & 40.15 & 90.33 & 34.32 &  91.87 & 26.14  & 92.98 & 26.40  & 93.25  & 85.23 \\

     & \cellcolor{gray!10}\emph{Ours} & \cellcolor{gray!10}\textbf{2.74}	& \cellcolor{gray!10}\textbf{98.82}	&  \cellcolor{gray!10}\textbf{24.32} & \cellcolor{gray!10}\textbf{93.73} &  \cellcolor{gray!10}\textbf{17.38} &  \cellcolor{gray!10}\textbf{95.65} & 	\cellcolor{gray!10}\textbf{18.58} & 	\cellcolor{gray!10}\textbf{95.38}& 	\cellcolor{gray!10}\textbf{17.01}& \cellcolor{gray!10}\textbf{95.89} & \cellcolor{gray!10}\textbf{86.40} \\
    \bottomrule
\end{tabular}
}
\vspace{0.2cm}
\caption{{\bf Quantitative results of OOD detection performance for ImageNet-1k as ID.} We employed our method on two pre-training paradigms (CLIP, and DINOv2). We use FPR95 and AUROC as evaluation metrics. We also report ID classification accuracy. The CLIP-based methods use ViT-B-16, and the DINOv2-based methods use ViT-B-14.}
% \vspace{-0.2cm}
\label{tab:in1k_comp}
\end{table*}
% \vspace{-0.2cm}
\section{Experiments}
\label{sec:exp}
% \vspace{-0.1cm}

In this section, we set up a benchmark for evaluating OOD performance in \cref{sec:Benchmark}. 
Then we compare our methods with the competitive baselines in \cref{sec:main_results}. 
We conduct ablation studies and present more analysis on our designed method in \cref{sec:analysis}.

% We first describe the evaluation datasets (Section~\ref{sec:dataset}) and  experimental setups (Section~\ref{sec:exp_setup}). In Section~\ref{sec:exp_results}, we show that MOS achieves state-of-the-art OOD detection performance, followed by extensive ablations that improve the understandings of MOS for large-scale OOD detection.% Section~\ref{sec:grouping_method_ablation}, Section~\ref{sec:class_num_ablation}, Section~\ref{sec:extractor_ablation}, and Section~\ref{sec:finetune_ablation}. %\SL{need section reference here}

\subsection{Benchmark}
\label{sec:Benchmark}
\noindent \textbf{In- and out-distribution Datasets.} To validate the effectiveness of our proposed method, we conduct evaluation on standard benchmarks, which use ImageNet-1K~\citep{data_imagenet} and ImageNet-100~\citep{ming2022delving} as the ID datasets.
%
% Following MCM~\citep{ming2022delving}, ImageNet-100 is constructed by randomly selecting 100 categories from ImageNet-1K~\citep{data_imagenet}. 
% %
% The selected classes are kept consistent with MCM~\citep{ming2022delving}. 
Following existing studies that leverage foundation models in OOD~\citep{Wang_2023_ICCV,li2024learning}, we use diverse OOD test datasets, including samples selected from iNaturalist18~\citep{van2018inaturalist}, SUN~\citep{xiao2010sun}, Places~\citep{zhou2017places}, and Textures~\citep{cimpoi2014describing}.

% \noindent \textbf{Out-of-distribution Datasets.} Following MOS~\citep{Huang_2021_CVPR}, we consider a diverse collection of OOD test datasets, including samples selected from iNaturalist~\citep{van2018inaturalist}, SUN~\citep{xiao2010sun}, Places~\citep{zhou2017places}, and Textures~\citep{cimpoi2014describing}. 
% %
% The details of OOD sample selection can be referred to MOS~\citep{Huang_2021_CVPR}.

% \noindent \textbf{Evaluation Metrics.}
% %
% For evaluation, we use the following metrics: (1) the false positive rate (FPR95) of OOD samples when the true positive rate of in-distribution samples is at 95\%, (2) the area under the receiver operating characteristic curve (AUROC), and (3) ID classification accuracy (ID ACC).

\noindent \textbf{Method Comparison.} We conduct method comparison on two pretaining paradigms(\ie CLIP and DINOv2). 
For each group, we apply some traditional scoring metric (such as MSP~\citep{hendrycks2016baseline}, MaxLogit~\citep{hendrycks2019scaling}, Energy~\citep{liu2020energy}, KNN~\citep{sun2022out}). 
Moreover, we also involve the current CLIP-based state-of-the-art methods, such as MCM~\citep{ming2022delving}, CLIPN~\citep{Wang_2023_ICCV}, NegPrompt~\citep{li2024learning}, and LSN~\citep{Nie_2024_ICLR}. We use KNN~\cite{sun2022out} as the scoring function when using DINOV2, and follow the scoring metric of CLIPN~\citep{Wang_2023_ICCV} when applying our method to CLIP. 
%
% More implementation details can be referred to supplementary material.

\begin{table*}[t]
    \centering
    \renewcommand\arraystretch{1.2}
    \resizebox{0.99\textwidth}{!}{
    \begin{tabular}{llccccccccccc}
        \toprule
              \multicolumn{2}{c}{\multirow{2.5}{*}{\textbf{Method}}} & \multicolumn{2}{c}{\textbf{iNaturalist18}} & \multicolumn{2}{c}{\textbf{Places}} & \multicolumn{2}{c}{\textbf{Sun}} & \multicolumn{2}{c}{\textbf{Textures}}  & \multicolumn{2}{c}{\bf Average} & \multirow{2.5}*{\bf ID ACC$\uparrow$}  \\
    \cmidrule(r){3-4} \cmidrule(r){5-6} \cmidrule(r){7-8} \cmidrule(r){9-10} \cmidrule(r){11-12}        
    \multicolumn{2}{c}{}& \bf FPR95$\downarrow$ &     \bf AUROC$\uparrow$ & \bf FPR95$\downarrow$ & \bf AUROC$\uparrow$ & \bf FPR95$\downarrow$ & \bf AUROC$\uparrow$ & \bf FPR95$\downarrow$ & \bf AUROC$\uparrow$ & \bf FPR95$\downarrow$ & \bf AUROC$\uparrow$   \\
    \midrule    
        \multirow{5}[1]{*}{\rotatebox{90}{CLIP-Based}} 
        & \makecell[l]{MSP  \citep{hendrycks2016baseline}} & 23.55 & 95.92  & 40.46 & 91.23 & 37.02 &92.45  &  24.40 & 94.90 & 31.43 & 93.63 & 91.93\\
        & \makecell[l]{MCM  \citep{ming2022delving}} & 18.13 & 96.77  & 34.52 & 94.36 & 36.45 &94.54  &  41.22 & 92.25 & 32.58 & 94.48 & 87.88\\
        & \makecell[l]{CLIPN  \citep{Wang_2023_ICCV}} & 4.87 & 98.16  & 13.64 & 96.93 & 13.55 & 97.56  & 15.78  & 93.02 & 11.96 & 96.41 & 91.64\\
        & \makecell[l]{LSN  \citep{Nie_2024_ICLR}} & 4.93 & 98.92  & 12.82 & 97.19 & 8.23 & 97.98  &  \bf8.26 & \bf 98.11 & 8.56 & 98.05 & 92.24\\
         & \cellcolor{gray!10}\emph{Ours} & \cellcolor{gray!10}\bf 3.20 & \cellcolor{gray!10}\bf 99.17 & \cellcolor{gray!10}\bf 10.05 & \cellcolor{gray!10}\bf 97.76 & \cellcolor{gray!10}\bf 7.06 & \cellcolor{gray!10}\bf 98.39 & \cellcolor{gray!10}9.31 & \cellcolor{gray!10}97.10 & \cellcolor{gray!10}\bf 7.40 & \cellcolor{gray!10}\bf 98.10 & \cellcolor{gray!10}\bf 92.85 \\
        \midrule
         \multirow{5}[1]{*}{\rotatebox{90}{Dinov2-Based}}
        & \makecell[l]{MSP  \citep{hendrycks2016baseline}} & 5.06 & 98.85 & 26.58  & 94.78 & 27.64 & 95.02 & 26.43 & 94.27 & 21.42 & 95.72 & 94.50\\
        & \makecell[l]{MaxLogit  \citep{hendrycks2019scaling}} & 5.55 & 98.76 & 29.69  &94.19 &32.73 & 94.20 & 29.27& 93.72 & 24.31 & 95.21 & 94.50\\
        & \makecell[l]{Energy  \citep{liu2020energy}} & 18.57 &96.69  &  54.72 & 88.92&62.42 & 87.17 & 57.28&  88.40& 48.21& 90.29 & 94.50\\
        & \makecell[l]{KNN  \citep{sun2022out}} & 2.58 & 99.02  & 18.45 & 95.12 & 15.89 & 96.16  & 16.79 & 96.38 & 13.42 & 96.66 & 94.50\\
        % &\makecell[l]{MOS~\citep{Huang_2021_CVPR}} &  &   &  &  &  &   &   &  &  &  & \\
         & \cellcolor{gray!10}\emph{Ours} & \cellcolor{gray!10}\textbf{2.25} & \cellcolor{gray!10}\textbf{99.23} & \cellcolor{gray!10}\textbf{12.81} & \cellcolor{gray!10}\textbf{96.66} & \cellcolor{gray!10}\textbf{8.51} & \cellcolor{gray!10}\textbf{97.86} & \cellcolor{gray!10}\textbf{8.85} & \cellcolor{gray!10}\textbf{97.28} & \cellcolor{gray!10}\textbf{8.10} & \cellcolor{gray!10}\textbf{97.75} & \cellcolor{gray!10}\textbf{96.94} \\
        \bottomrule
    \end{tabular}}
    \vspace{0.2cm}
\caption{{\bf Quantitative results of OOD detection performance for ImageNet-100 as ID.} We employed our method on two pre-training paradigms (CLIP, and DINOv2). We use FPR95 and AUROC as evaluation metrics. We also report ID classification accuracy. The CLIP-based methods use ViT-B-16, and the DINOv2-based methods use ViT-B-14.} 
\label{tab:in100_comp}
% \vspace{0.4cm}
\end{table*}

\subsection{Main Results}
\label{sec:main_results}

% \vspace{-1cm}

\noindent \textbf{Results on ImageNet-1K.} We compare the proposed approach with the state-of-the-art methods for ImageNet-1K as ID on \cref{tab:in1k_comp}.
These results show: 1) Based on DINOv2, our method reaches the best performance when setting ImageNet-1K as ID. 
Specifically, our approach reaches 17.01\% FPR95 and 95.89\% AUROC, averaging the results of all the OOD test sets. 
Our method surpasses MOS~\citep{Huang_2021_CVPR} by 9.39\% in FPR95, and 2.64\% in AUROC, which proves the importance of learning expert models for OOD detection.
2) When applying our method to CLIP, our method reaches 20.02\% and 94.89\%, which also outperforms NegPrompt~\cite{li2024learning} by a large margin. 
These results indicate the effectiveness of the proposed MoFE and the dynamic regularized  Mixup.
3) Our approach reaches 2.74\% FPR95 on iNaturalist18 and increases the performance on all the test sets, which indicates that our MoFE design retains the discriminativeness of DINOv2 and facilitates feature learning on various feature subspaces.

\noindent \textbf{Results on ImageNet-100.}
We compare the proposed approach with the state-of-the-art methods for ImageNet-100 as ID on \cref{tab:in100_comp}.
Based on DINOv2, our method reaches 8.10\% FPR95 and 97.75\% AUROC, surpassing the baseline by 4.40\% FPR95, and 0.23\% AUROC.
This indicates that our proposed approach is also effective in a small-scale ID dataset. 
On the other hand, when applying our method to CLIP, we achieve 7.40\% FPR95 and 98.10\% AUROC, outperforming LSN~\cite{Nie_2024_ICLR} by 1.16\% FPR95.  
The above experimental results validate the effectiveness of our approach, and we achieve the best performance on both small-scale and large-scale ID datasets. 

% %
% Note that the improvement on ImageNet-100 is not noticeable compared to ImageNet-1K. 
% %
% The reason might be that MoE tends to be effective on large-scale ID datasets, a point we will also verify in our subsequent analysis.

% \vspace{-0.3cm}

\subsection{Analysis}

\label{sec:analysis}

In this section, we conduct more analysis on the proposed methods. 
We use the DINOv2 as the pretaining paradigm and KNN~\cite{sun2022out} as the scoring function.
We use ImageNet-1K as ID data and report the average performance on the four out-of-distribution datasets mentioned in \cref{sec:Benchmark}. 
The baseline is set to DINOv2 with KNN scoring function without our designed method.

\noindent \textbf{Contributions of Individual Components.} As shown in \cref{tbl:ablation}, we evaluate the contribution of each component of the full method.
%
% The baseline is set to DINOv2 with KNN scoring function without our designed method.  
%
On ImageNet-1K, MoFE and Dynamic-$\beta$ Mixup contribute 6.68\% and 5.41\% FPR95, respectively.
When combined, the best performance (17.01\% FPR95, and 95.89\% AUROC) is achieved.
This validates our design consideration in that they are complementary and should be combined.

%
% While on ImageNet-100, most of the performance gain is achieved by DRM, which reaches 8.10\% FPR95, and 97.75\% AUROC.
%
% The primary reason is that, in large-scale ID datasets, the interference between data during finetuning is stronger. 
% %
% Therefore, MoE tends to be more effectiveness in large-scale ID datasets. 

\begin{table}[t]
\small
\centering
\renewcommand{\arraystretch}{1.05}
\vspace{-0.7cm}
\resizebox{1.0\linewidth}{!}{ 
\begin{tabular}{lcccc}
    \toprule
    \textbf{Settings}  & Baseline & + MoFE & + D-$\beta$ & + MoFE+D-$\beta$ \\
    \midrule
    \bf FPR95$\downarrow$ & 29.27 & 22.59 & 23.85 & \bf 17.01  \\
    \bf AUROC$\uparrow$ & 92.67 & 94.01 & 93.72 & \bf 95.89  \\
    \bottomrule
\end{tabular}}
% \vspace{0.1cm}
\caption{Ablation study of individual components for ImageNet-1k as In-Distribution dataset.}
\label{tbl:ablation}
\end{table}

\begin{table}[t]
\small
\centering
\renewcommand{\arraystretch}{1.0}
\vspace{-0.5cm}
\resizebox{1.0\linewidth}{!}{ 
\begin{tabular}{l>{\centering\arraybackslash}p{0.8cm}>{\centering\arraybackslash}p{0.8cm}>{\centering\arraybackslash}p{0.8cm}>{\centering\arraybackslash}p{0.8cm}>{\centering\arraybackslash}p{0.8cm}>{\centering\arraybackslash}p{0.8cm}>{\centering\arraybackslash}p{0.8cm}}
    \toprule
    \textbf{Num.}  & 2 & 3 & 5 & 7 & 8 & 9 \\
    \midrule
    \textbf{Gain} & 4.10 & 6.30 & 9.80 & 12.26 & 12.10 & 12.09 \\
    \bottomrule
\end{tabular}}
% \vspace{0.1cm}
\caption{The effect of Cluster Number. We report the performance gain in FPR95 compared to the model without MoFE.}
\label{fig:cluster_num}
\end{table}

\noindent \textbf{Cluster Number.}
We conduct an experiment to validate the impact of cluster number on MoFE performance. 
We set different numbers of clusters. 
As shown in \cref{fig:cluster_num}, we report the performance gain in FPR95. 
As the increasing of cluster number, the performance gradually increases. 
The performance saturates when the cluster number reaches 7.

\noindent \textbf{Feature Space Separation.}
As shown in \cref{tbl:Grouping}, we validate the different strategies for determining the subspace. 
We compare our method with two methods: WordNet and clustering. 
The results show that using the ours (i.e. WordNet + Clustering) is the most promising approach. 
The reason might be that the features extracted by pretrained model are discriminative, especially at coarse-grained level. 
Therefore, the feature similarity can be used to refine the initial cluster from WordNet. 

\noindent \textbf{More Analysis on Dynamic-$\beta$ Mixup.}
As shown in \cref{tbl:DRM}, we conduct an ablation study on Dynamic-$\beta$ Mixup. 
When we remove the regularization term, we find that the performance degrades (30.43\% FPR95).  
Moreover, when we dynamic beta distribution is removed, the performance decreases to 24.96\% FPR95. 
These results validate the importance of both components in Dynamic-$\beta$ Mixup.

\begin{table}[t]
\small
\centering
\renewcommand{\arraystretch}{1.0}
\vspace{-0.7cm}
\resizebox{1.0\linewidth}{!}{ 
\begin{tabular}{lcccc}
    \toprule
    \textbf{Grouping}  & Baseline & WordNet & Clustering & Ours \\
    \midrule
    \bf FPR95$\downarrow$ & 29.27 & 25.63 & 26.34 & \bf 22.59  \\
    \bf AUROC$\uparrow$ & 92.67 & 93.01 & 92.99 & \bf 94.01  \\
    \bottomrule
\end{tabular}}
% \vspace{0.1cm}
\caption{Analysis on Grouping Strategy in MoFE.}
\label{tbl:Grouping}
\end{table}

\begin{table}[t]
\small
\centering
\renewcommand{\arraystretch}{1.0}
\vspace{-0.2cm}
\resizebox{1.0\linewidth}{!}{ 
\begin{tabular}{l>{\centering\arraybackslash}p{1.1cm}>{\centering\arraybackslash}p{1.3cm}>{\centering\arraybackslash}p{1.3cm}>{\centering\arraybackslash}p{1.3cm}>{\centering\arraybackslash}p{1.3cm}}
    \toprule
    \textbf{Methods}  & Baseline & w/o Reg & w/o D-$\beta$ & Ours\\
    \midrule
    \bf FPR95$\downarrow$ & 29.27 & 30.43 & 24.96 & \bf 23.85  \\
    \bf AUROC$\uparrow$ & 92.67 & 91.65 & 93.36 & \bf 93.72  \\
    \bottomrule
\end{tabular}}
% \vspace{0.1cm}
\caption{Ablation study of Dynamic-$\beta$ Mixup.}
\label{tbl:DRM}
\end{table}

\section{Conclusion}

This paper studies the OOD detection task within the context of foundation models.
Our study shows that vision foundation models (e.g., DINOv2) are effective OOD detectors, suggesting high-quality and generalizable feature space is essential for OOD detection. 
Our study highlights that CLIP's pre-trained feature space is less effective for fine-grained tasks, where DINOv2 performs significantly better, which worths further exploration. 
Second, we find that simply fine-tuning foundation models on ID data will result in performance degradation due to the loss of generalization ability. 
In order to further optimize the performance of OOD detection when ID data is available for fine-tuning, we propose MoFE and a Dynamic-$\beta$ Mixup data augmentation to enhance the feature learning.
We conduct extensive experiments and ablation studies to validate the effectiveness of our approach.
We believe enhancing the discriminativeness and generalization ability of learned features is the key to OOD detection.
We hope our investigation could inspire more future studies.

\clearpage
{
    \small
    \bibliographystyle{ieeenat_fullname}
    \bibliography{main}
}

% \clearpage

\clearpage
\setcounter{page}{1}

\appendix 
\begin{table*}[]
\centering
\renewcommand\arraystretch{1.4}
\resizebox{0.99\textwidth}{!}{
\begin{tabular}{llccccccccccc}
    \toprule
              \multicolumn{2}{c}{\multirow{2.5}{*}{\textbf{Method}}} & \multicolumn{2}{c}{\textbf{iNaturalist18}} & \multicolumn{2}{c}{\textbf{Places}} & \multicolumn{2}{c}{\textbf{Sun}} & \multicolumn{2}{c}{\textbf{Textures}}  & \multicolumn{2}{c}{\bf Average} & \multirow{2.5}*{\bf ID ACC$\uparrow$}  \\
\cmidrule{3-12}          & & FPR95$\downarrow$ & AUROC$\uparrow$ & FPR95$\downarrow$ & AUROC$\uparrow$ & FPR95$\downarrow$ & AUROC$\uparrow$ & FPR95$\downarrow$ & AUROC$\uparrow$ & FPR95$\downarrow$ & AUROC$\uparrow$   \\
\midrule 
\multirow{5}[2]{*}{\rotatebox{90}{\makecell[c]{IN-1K Pretrained \\( SoTA )}}} 
& \makecell[l]{Energy~\citep{liu2020energy}} & 55.72 & 89.95 & 59.26 & 85.89 & 64.92 & 82.86 & 53.72 & 85.99 & 58.41 & 86.17 & 75.08 \\
& \makecell[l]{MSP~\citep{hendrycks2016baseline}} & 54.99 & 87.74 & 70.83 & 80.86 & 73.99 & 79.76 & 68.00 & 79.61 & 66.95 & 81.99 & 75.08 \\
& \makecell[l]{MaxLogit~\citep{hendrycks2019scaling}} & 54.05 & 87.43 & 72.98 & 78.03 & 73.37 & 78.03 & 68.85  & 79.06 &  67.31 & 80.64  & 75.08\\
& \makecell[l]{KNN~\citep{sun2022out}} & 7.30 &  98.46 & 48.40 &  88.24  & 56.46 & 88.14 &  39.91 & 89.23 & 38.02  &  91.01  &75.08 \\
& \makecell[l]{MOS~\citep{Huang_2021_CVPR}} & 9.54 &  98.23 & 43.62 & 91.26 & 48.15 & 90.42    & 57.12 &  83.16 &  39.60 & 90.76 & 75.20\\
\midrule
\multirow{6}[1]{*}{\rotatebox{90}{\makecell[c]{CLIP-Based }}} 
& \makecell[l]{Energy \citep{liu2020energy}} &  65.00 &  87.17 & 57.40 & 87.32 & 46.43  & 91.17 & 57.40  & 87.32 &  56.55 & 88.24  & 79.39 \\
& \makecell[l]{MSP \citep{hendrycks2016baseline}} &40.89& 88.63& 65.81& 81.24& 67.90& 80.14& 64.96& 78.16& 59.89& 82.04 & 79.39 \\
& \makecell[l]{MaxLogit \citep{hendrycks2019scaling}} & 60.86 & 88.03 & 55.5 & 87.44 &  44.81& 91.16 & 52.25  & 86.04 &  53.35 & 88.16  & 79.39\\
& \makecell[l]{MCM  \citep{ming2022delving}}
& 30.91 & 94.61 & 37.59&  92.57&  44.69&  89.77&  57.77&  86.11&  42.74&  90.77 & 67.01 \\
& \makecell[l]{CLIPN  \citep{Wang_2023_ICCV}}  
& 23.94 & 95.27 & 26.17 &  93.93 &  33.45 &  92.28 &  40.83 &  90.93&  31.10&  93.10 & 68.53 \\
& \makecell[l]{LSN  \citep{Nie_2024_ICLR}} & 21.56 & 95.83  & 34.48 & 91.25 & 26.32 & 94.35  &  38.54 & 90.42 & 30.22 & 92.96 & 71.89\\
\midrule
\multirow{4}[2]{*}{\rotatebox{90}{\makecell[c]{Dinov2-Based }}} 
& \makecell[l]{Energy  \citep{liu2020energy}} & 13.23 &  96.86 &  66.63 & 83.32 & 61.57  & 84.76 & 66.43  & 82.36 &  51.96  &  86.82  & 81.70\\
& \makecell[l]{MSP  \citep{hendrycks2016baseline}} & 9.05 & 98.15 & 52.58 & 86.34 & 49.45 & 87.35 & 52.32  & 85.82 & 40.85 & 89.41  & 81.70\\
& \makecell[l]{MaxLogit  \citep{hendrycks2019scaling}} & 8.21  &  98.22 & 53.93 & 85.80 & 50.48 & 87.00 & 54.32  & 85.25 & 41.73 & 89.06  & 81.70\\
& \cellcolor{gray!10}{KNN}~\citep{sun2022out} & \cellcolor{gray!10}{3.01} & \cellcolor{gray!10}{98.26}  & \cellcolor{gray!10}{42.78} & \cellcolor{gray!10}{88.89} & \cellcolor{gray!10}{35.96} & \cellcolor{gray!10}{91.51}  &  \cellcolor{gray!10}{35.30} & \cellcolor{gray!10}{91.05} & \cellcolor{gray!10}{29.27} & \cellcolor{gray!10}{92.67} &\cellcolor{gray!10}{81.70}\\
\midrule
& \cellcolor{gray!10}{Naive finetuning} & \cellcolor{gray!10}{5.67} & \cellcolor{gray!10}{97.65}  & \cellcolor{gray!10}{43.25} & \cellcolor{gray!10}{88.21} & \cellcolor{gray!10}{36.42} & \cellcolor{gray!10}{90.21}  &  \cellcolor{gray!10}{28.04} & \cellcolor{gray!10}{92.66} & \cellcolor{gray!10}{28.34} & \cellcolor{gray!10}{92.18} & \cellcolor{gray!10}{85.96}\\
\bottomrule
\end{tabular}
% \label{tab:pilot_comparision} 
}
\caption{{\bf Quantitative results of OOD detection performance for ImageNet-1k as ID.} We conduct three pre-training paradigms (ImageNet Pretrained (IN-1K), CLIP, and DINOv2) for comparison. We use FPR95 and AUROC as evaluation metrics. We also report ID classification accuracy.}
\label{tab:hie_method}
\end{table*}

\section{Appendix}

\subsection{Implementation Details}
We adopt ViT-Base~\citep{dosovitskiy2021vit} as the backbone. 
When using pre-training paradigms of CLIP and DINOv2, we directly initialize ViT from their weights. 
Besides, when using CLIP, we leverage CLIPN~\citep{Wang_2023_ICCV} as the baseline method and we follow their scoring metric. 
For DINOv2, we use DINOv2 with standard cross-entropy loss as the baseline method and the scoring metric is KNN~\citep{sun2022out}. 
When using DINOv2, we first conduct linear probing for 3 epoches to ensure its training stability. 
Our models are trained with AdamW optimizer with $\beta_s= \{0.9, 0.95\}$, with an effective batch size of 1024 on 8 NVIDIA 3090 GPUs.
The values for weight decay and layer decay are 0.05 and 0.75, 
The training epochs are set to 40.
We set a cosine learning rate schedule and the minimum learning rate is 1e-6.

\subsection{Implementation Details of Naive Finetuning}
\label{sec:Naive_finetuning}

The model is trained with cross entropy loss and Adam optimizer with $\beta_s= \{0.9, 0.95\}$, with an effective batch size of 1024 on 8 NVIDIA 3090 GPUs. The values for weight decay and layer decay are 0.05 and 0.75. The training epochs are set to 40. We set a cosine learning rate schedule, and the minimum learning rate is 1e-6. We first conduct linear probing for 3 epochs to ensure their training stability. During the testing phase, we use KNN as the classifier using features from the penultimate layer.

% \subsection{Hierarchical Classification based Methods}
% %
% We notice that these studies~\cite{Li_2023_ICCV, fginference, Huang_2021_CVPR} are relevant to ours, which also break the semantic space into smaller ones. 
% %
% They first classify among the superclasses, and then classifies fine-grained subclasses and conduct OOD detection within the identified superclass. 
% %
% Different from these approaches, we build an effective MoFE framework and utilize multiple experts with each expert specializing in a specific subspace. 
% %
% Our approach facilitates learning more subspace-specific discriminative features and preventing interference between features from different partitions. 
% %
% As shown in \cref{tab:hie_method}, we select the most effective method HVCM~\cite{Li_2023_ICCV} for comparison and we implement it based on DINOv2. 
% %
% The results show that our approach significantly surpasses it by a large margin, which reveals the importance of learning expert models for OOD tasks. 

\subsection{Comparison with the traditional MoE} 

The proposed Mixture of Feature Expert (MoFE) is specifically tailored for OOD detection with foundation models, which is different from the original MoE designed for general LLM and vision tasks from both insights and methods. 
In terms of insights,  our MoFE was crafted to reduces the difficulty of fitting complex data distribution when training foundation models on limited In-Distribution (ID) data, while MoE is initially designed to accelerate inference for large models~\cite{vmoe} and is leveraged for learning visual attributes for domain generalization~\cite{sparse_MOE}.  
We're not aware of any existing work that shares our insights. 
In terms of method design,  as our primary insights are to prevent features from collapsing to the ID data distribution, we partition the feature space into different subspaces and design routing mechanism based on feature similarities.  
Our routing mechanism leverages the class token, which contains the most discriminative feature, to guide all the features to the specific expert. 
%
% More importantly, this allows us to enhance ID classification performance and OOD detection performance when tuned on ID data. 

\subsection{Further Evaluation for Pilot Study.}

\label{sec:more_pilot}

% \subsubsection{Further Evaluation for Pilot Study}
%
We conduct further validation for pilot study, where we select data from OpenImage~\cite{openimages} for experiments. 
Specifically, we randomly select 1000 classes as the ID data. 
Furthermore, we randomly sample another 1000 classes as the OOD data, which is denoted as subset 1. 
For constructing a finegrained OOD subset, we select the categories which are closely related to the ID categories, where semantically belong to the same superclasses with the ID data according to WordNet. We denote it as subset 2. 
The results in ~\cref{tab:further_pilot} demonstrate that 1) Foundation models surpass the ImageNet pretrained methods by a large margin. 2) DINOv2 performs better than CLIP in the finegrained OOD tasks. 
For example, DINOV2 with KNN achieve 17.23\% FPR95 in subset 2, while the CLIP based method can only achieve 29.87\% FPR95. 

\subsection{Limitation}
We summarize the limitations of our research as follows:
Although CLIP and DINOv2 are currently the top foundation models, they still have inherent shortcomings. For instance, CLIP only utilizes image-text pairs for contrastive learning between text and images, lacking self-supervised learning on images. This results in its inability to capture fine-grained image details, leading to poor performance on finegrained tasks. On the other hand, DINOv2 employs a large number of images for self-supervised learning, yet it still performs poorly on certain categories, indicating potential long-tail distribution issues in its pre-training data.
The current benchmarks for OOD detection have significant limitations. While they utilize datasets like ImageNet-1K, which cover a wide range of categories, the OOD data itself is relatively limited.

\begin{table*}[t]
\label{tab:in1k}
\centering
\renewcommand\arraystretch{1.2}
\resizebox{0.99\textwidth}{!}{
\begin{tabular}{llccccccc}
    \toprule
              \multicolumn{2}{c}{\multirow{2.5}{*}{\textbf{Method}}} & \multicolumn{2}{c}{\textbf{Subset 1}} & \multicolumn{2}{c}{\textbf{Subset 2}}  & \multicolumn{2}{c}{\bf Average} & \multirow{2.5}*{\bf ID ACC$\uparrow$}  \\
\cmidrule{3-8}          & & FPR95$\downarrow$ & AUROC$\uparrow$ & FPR95$\downarrow$ & AUROC$\uparrow$  & AUROC$\uparrow$ & FPR95$\downarrow$ & AUROC$\uparrow$   \\
\midrule 
\multirow{5}[2]{*}{\rotatebox{90}{\makecell[c]{IN-1K Pretrained \\( SoTA )}}} 
& \makecell[l]{Energy~\citep{liu2020energy}}  & 60.23 & 76.23 & 74.66 & 73.21 & 67.44 & 74.71 & 72.33 \\
& \makecell[l]{MSP~\citep{hendrycks2016baseline}}  & 58.23 & 79.01 & 72.41 & 77.23 & 65.32 & 78.12 & 72.33  \\
& \makecell[l]{MaxLogit~\citep{hendrycks2019scaling}}  & 57.35 & 79.32 & 70.23  & 78.33 & 63.79  & 78.82  & 72.33  \\
& \makecell[l]{KNN~\citep{sun2022out}} & 15.01 & 96.55 &  33.24 & 94.01 & 24.12  &  95.28  &  72.33 \\
& \makecell[l]{MOS~\citep{Huang_2021_CVPR}}  & 17.37 & 97.01    & 35.44 &  93.26 & 26.41  & 95.14 & 73.46\\
\midrule
\multirow{6}[1]{*}{\rotatebox{90}{\makecell[c]{CLIP-Based }}} 
& \makecell[l]{Energy \citep{liu2020energy}} & 57.43  & 92.88 & 65.12  & 79.23 & 61.27  &  86.10 & 78.64 \\
& \makecell[l]{MSP \citep{hendrycks2016baseline}} & 43.23& 89.88& 62.21& 79.11& 52.72 &  84.50 &  78.64 \\
& \makecell[l]{MaxLogit \citep{hendrycks2019scaling}} &  45.87& 90.16 & 60.23  & 80.12 & 53.04  & 85.14  & 78.64\\
& \makecell[l]{MCM  \citep{ming2022delving}}
&  23.34&  94.41&  45.01&  92.16& 34.17 &  93.28 & 65.27 \\
& \makecell[l]{CLIPN  \citep{Wang_2023_ICCV}}  
 &  10.14 &  96.88 &  30.21 &  94.01& 20.18 &  95.44 & 64.34\\
& \makecell[l]{LSN  \citep{Nie_2024_ICLR}}  & 9.87 & 95.12  &  29.87 & 95.76 & 19.87  & 95.43 &   72.81\\
\midrule
\multirow{4}[2]{*}{\rotatebox{90}{\makecell[c]{Dinov2-Based }}} 
& \makecell[l]{Energy  \citep{liu2020energy}} & 50.23  & 88.23 & 64.13  & 83.21 &  57.18  &  85.71  & 82.41\\
& \makecell[l]{MSP  \citep{hendrycks2016baseline}} & 31.38 & 93.98 & 54.32  & 86.98 & 42.85&  90.48 & 82.41\\
& \makecell[l]{MaxLogit  \citep{hendrycks2019scaling}} & 30.23 & 94.02 & 56.32  & 86.45 & 43.27 & 90.23  & 82.41\\
& \cellcolor{gray!10}{KNN~\citep{sun2022out}}&  \cellcolor{gray!10}\textbf{8.16} &  \cellcolor{gray!10}\textbf{97.26}  &   \cellcolor{gray!10}\textbf{17.23} &  \cellcolor{gray!10}\textbf{96.38} &  \cellcolor{gray!10}\textbf{12.70} &  \cellcolor{gray!10}\textbf{96.82} & \cellcolor{gray!10}\textbf{82.41} \\
\midrule
% & Naive finetuning & 5.67 & 97.65  & 43.25 & 88.21 & 36.42 & 90.21  &  28.04 & 92.66 & 28.34 & 92.18 & 85.96\\
% \bottomrule
\end{tabular}
}
\caption{{\bf Pilot Study using data from OpenImage~\cite{openimages}.} We conduct three pre-training paradigms (ImageNet-1K (IN-1K) Pretrained, CLIP, and DINOv2) for comparison. We use FPR95 and AUROC as evaluation metrics. We also report ID classification accuracy.}
\label{tab:further_pilot} 
\end{table*}

\begin{figure*}[h]
    \centering
    \subfloat[Train with MOS]{
        \includegraphics[width=0.4\textwidth]{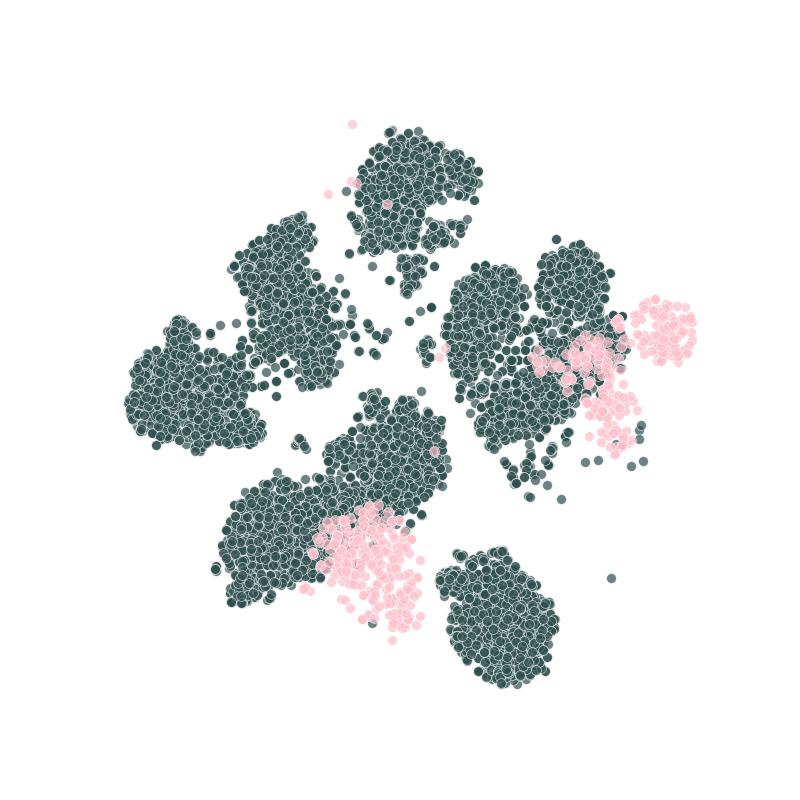}
    }
    % \vspace{-0.3cm}
    % \subfloat[Train without MoFE]{
    %     \includegraphics[width=0.85\textwidth]{ICLR 2025 Template/figures/evidence_figures/fig1_sub2.pdf}
    % }
    % \vspace{-0.3cm}
    \subfloat[Train with MoFE]{
        \includegraphics[width=0.4\textwidth]{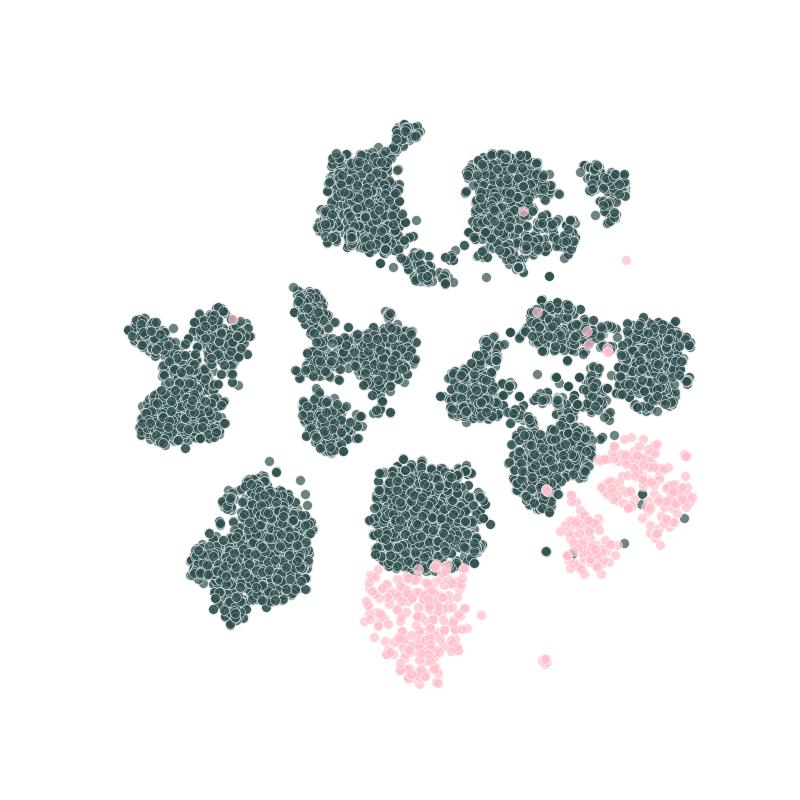}
    }
    % \vspace{-0.1cm}
    \caption{\textbf{Visualization of feature space of MoFE and MOS.} It can be observe that, trained with MOS, the outlier features are still mingled with in-domain data, while MoFE can well separate the in- and out-of-distribution data.}
    \label{tab:motivation_mofe}
    % \vspace{-0cm}
        % \vspace{-0.3cm}
\end{figure*}

\begin{figure*}[h]
    \centering
    \subfloat[Train with Mixup]{
        \includegraphics[width=0.40\textwidth]{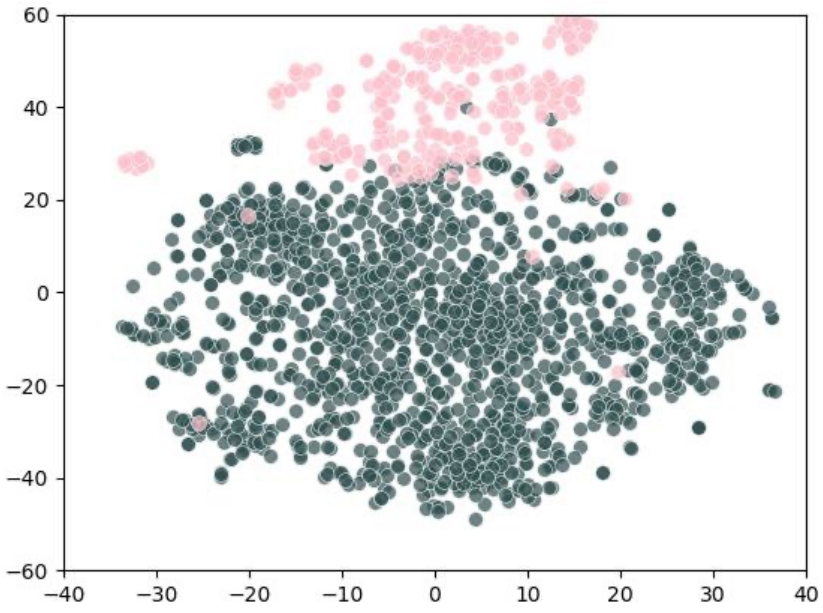}
    }
    \hfill
    \subfloat[Train without Mixup]{
        \includegraphics[width=0.40\textwidth]{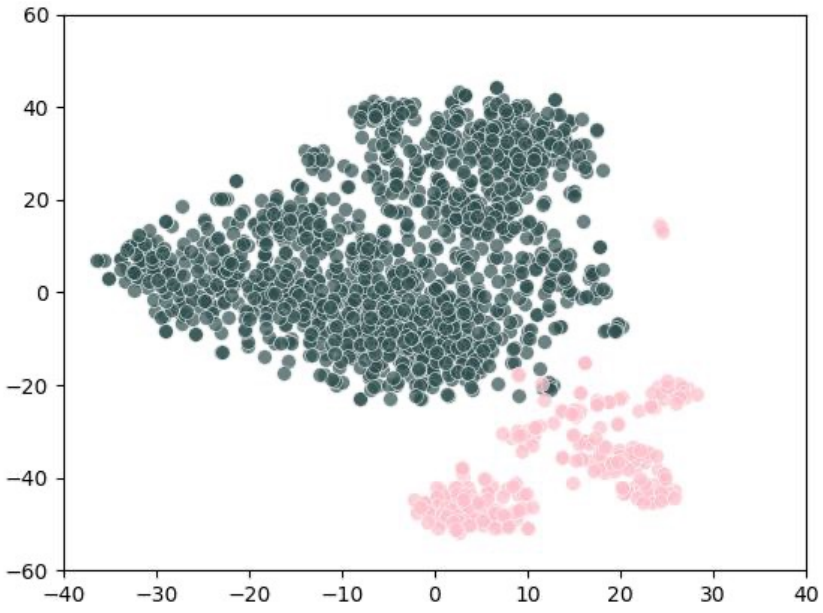}
    }
    \vspace{-0.1cm}
    \caption{\textbf{The effect of vanilla mixup on the feature space of DINOv2.} We can observe that vanilla Mixup can blur the decision boundary between ID and OOD.}
    % \vspace{-5cm}
    \label{fig:motivation_mixup}
            \vspace{-0.2cm}
\end{figure*}

\begin{figure*}[t]
    \centering
    \subfloat[CLIP - coarse-grained]{
        \includegraphics[width=0.3\textwidth]{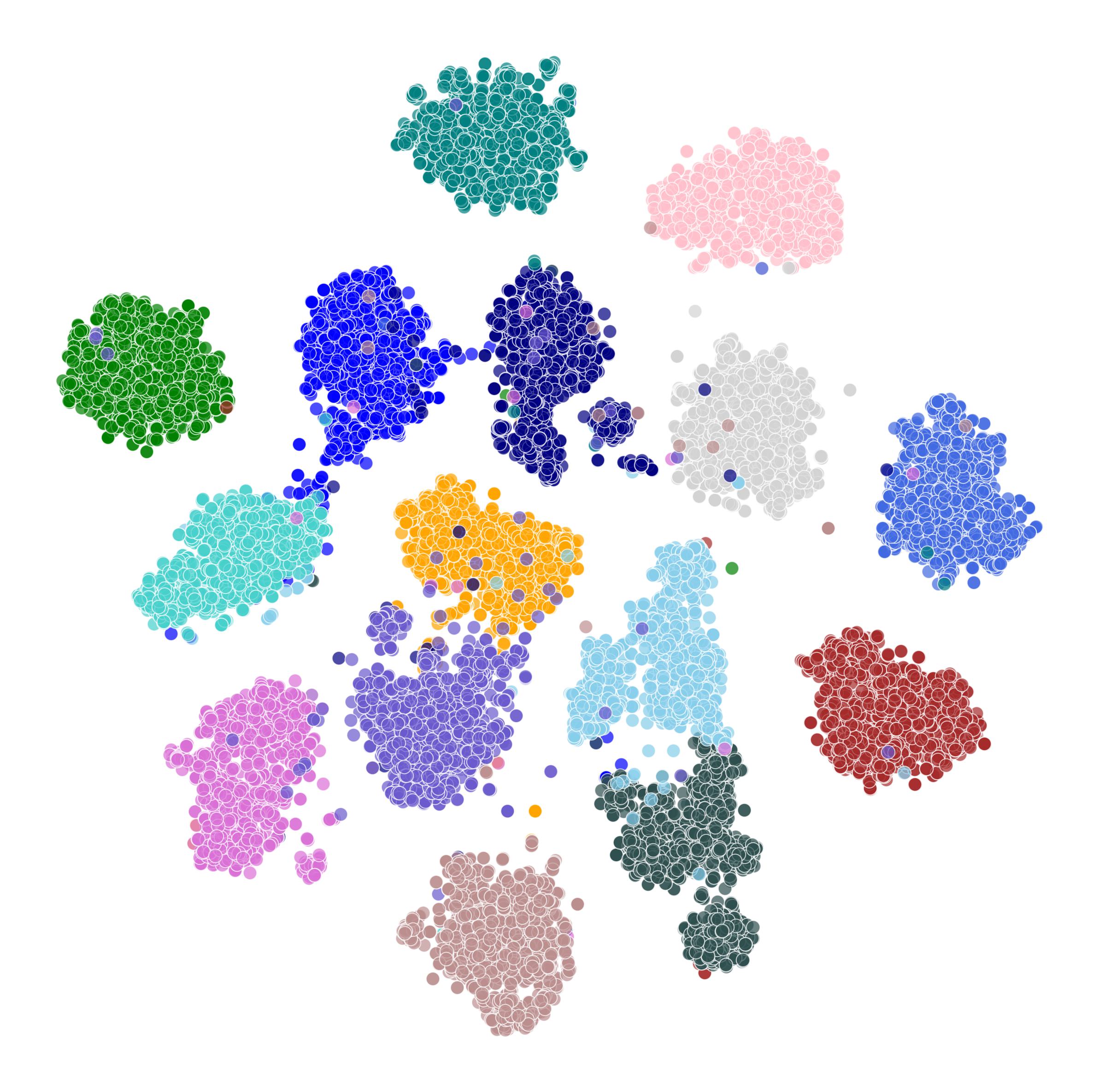}
        \label{fig:clip_vast}
    }
    \hfill
    \subfloat[CLIP - fine-grained]{
        \includegraphics[width=0.3\textwidth]{figures/figure1_feature_space_v4/clip_finegrained.jpg}
        \label{fig:clip_finegrain}
    }
    \hfill
    \subfloat[CLIP - failure]{
        \includegraphics[width=0.3\textwidth]{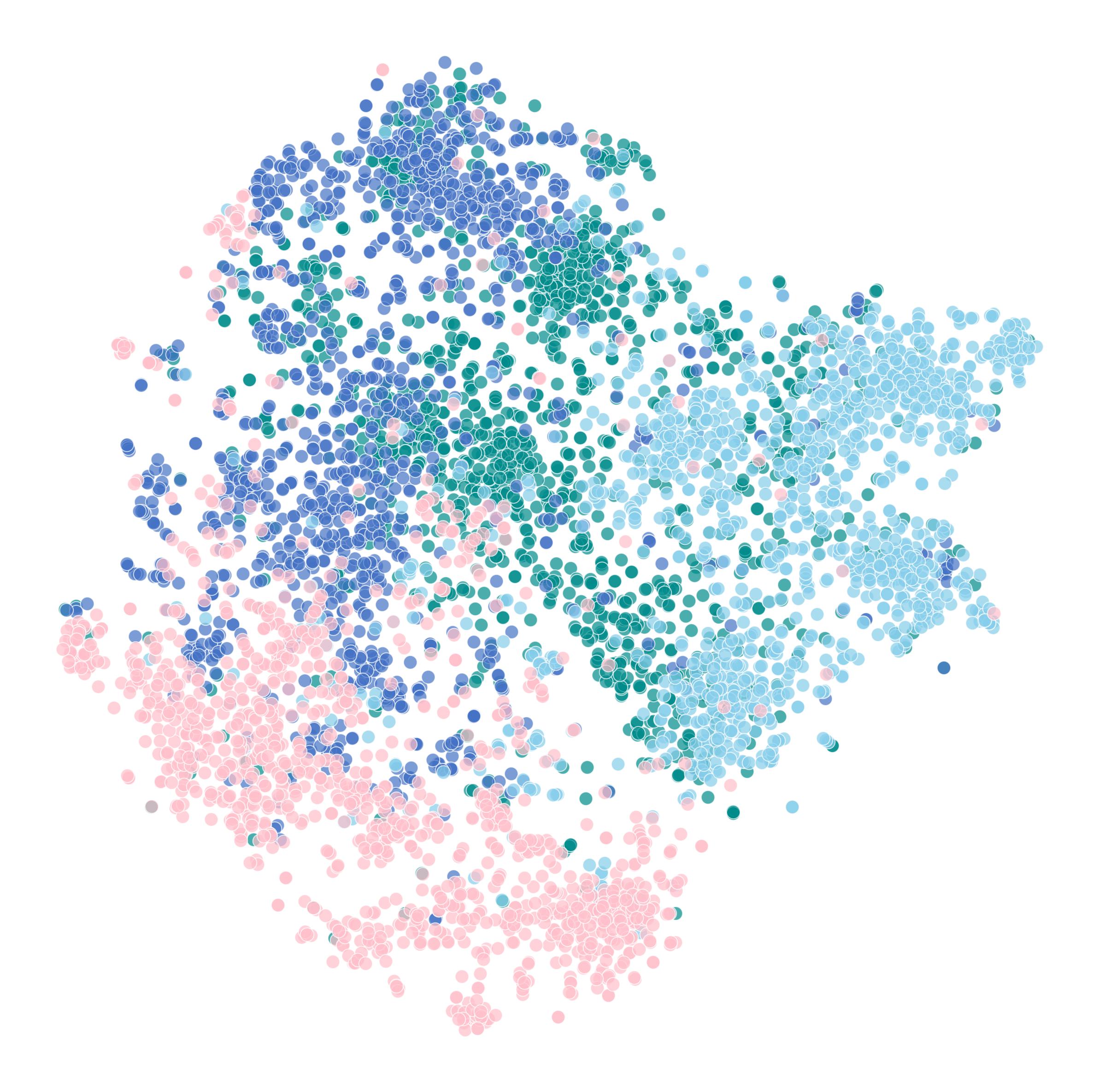}
        \label{fig:clip_failure}
    }
    \vspace{-0.2cm}
    \subfloat[DINOv2 - coarse-grained]{
        \includegraphics[width=0.3\textwidth]{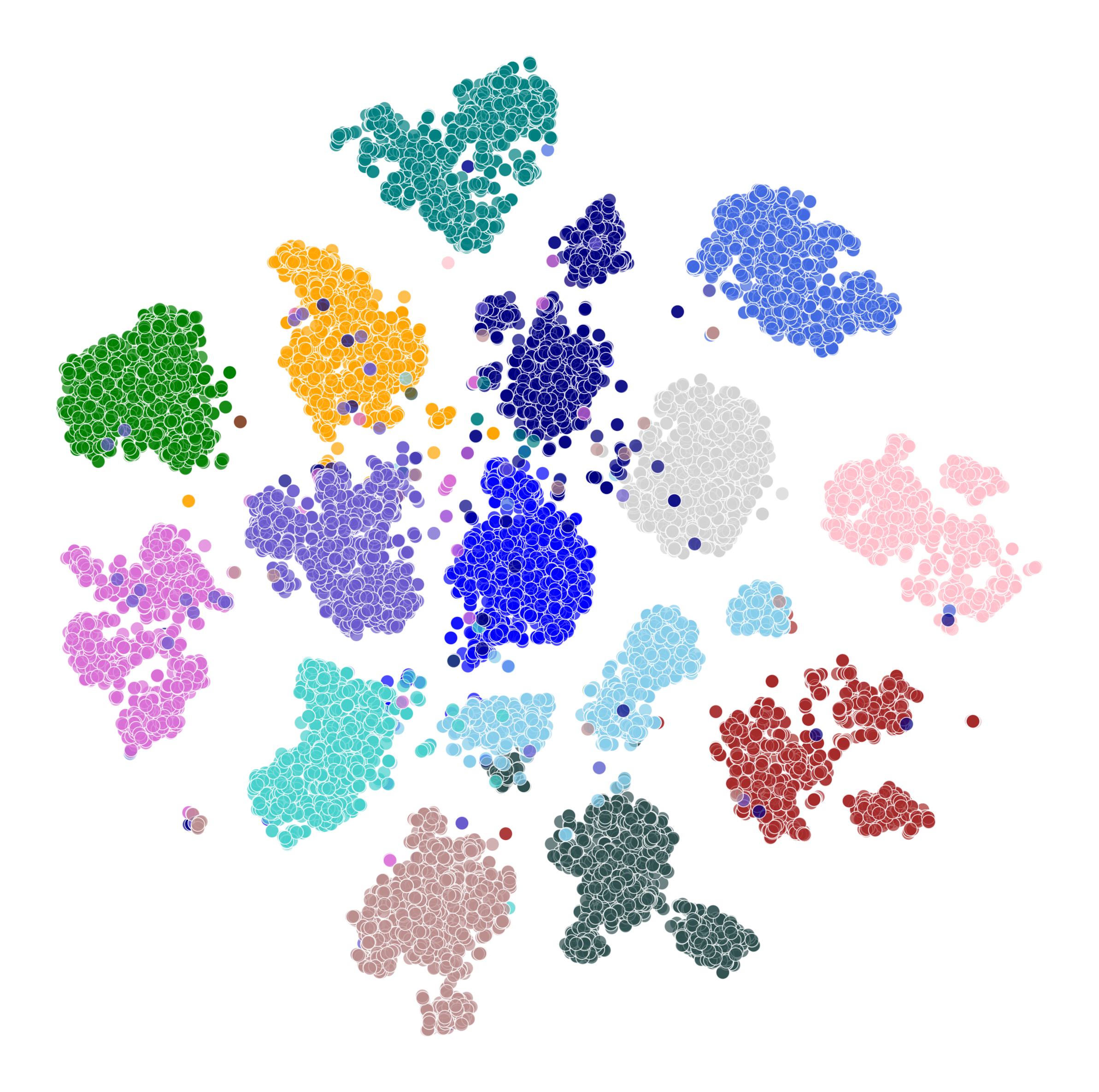}
        \label{fig:dinov2_vast}
    }
    \hfill
    \subfloat[DINOv2 - fine-grained]{
        \includegraphics[width=0.3\textwidth]{figures/figure1_feature_space_v4/dinov2_finegrained.jpg}
        \label{fig:dinov2_finegrain}
    }
    \hfill
    \subfloat[DINOv2 - failure]{\includegraphics[width=0.3\textwidth]{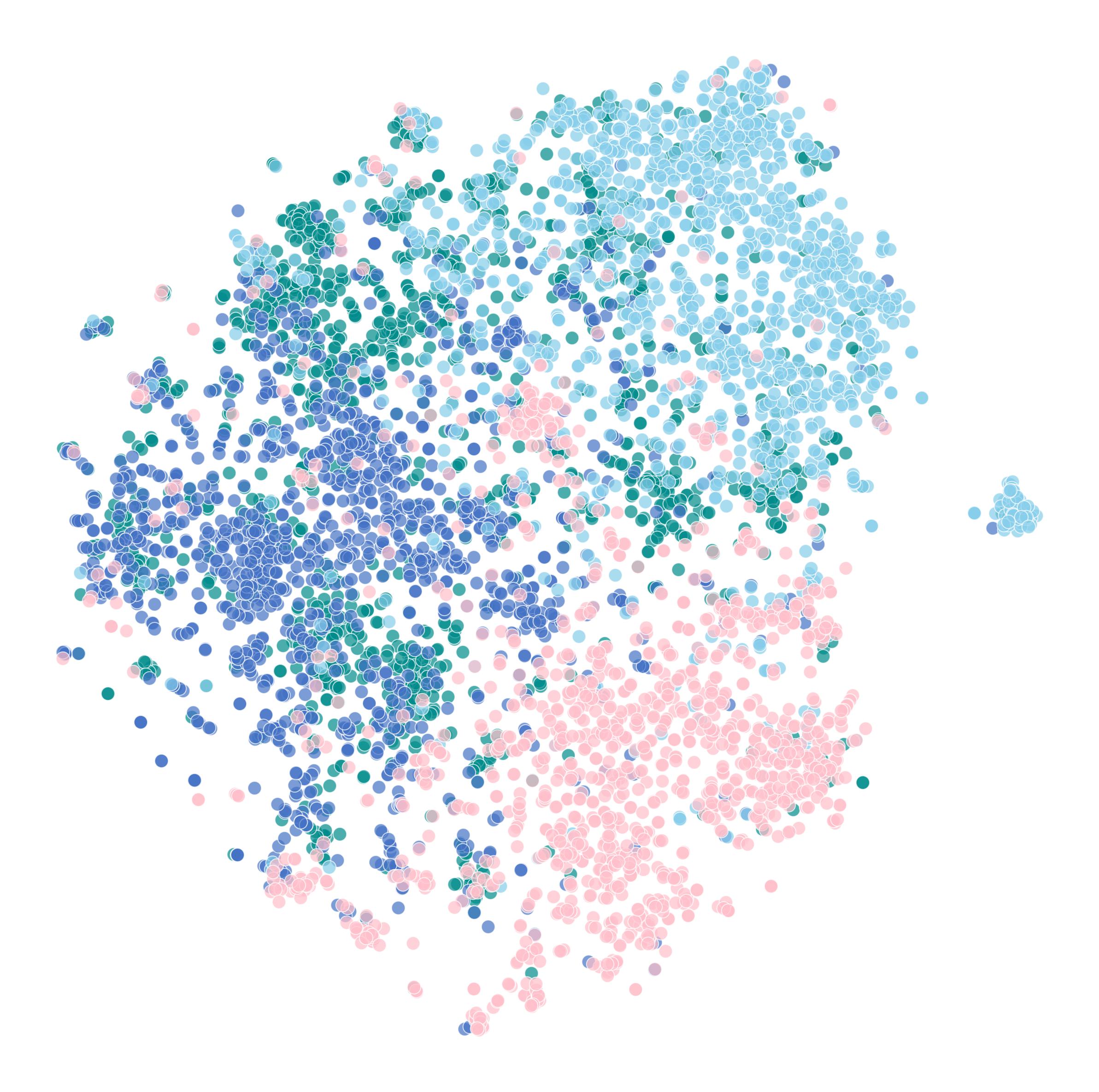}
        \label{fig:dinov2_failure}
    }
    \caption{{\bf Feature Space Visualization for Foundation Models.} The first row shows the feature space for CLIP and the second is for DINOv2. For each of them, we visualize the features of coarse-grained categories, fine-grained categories, and some failure cases.
For the coarse-grained feature visualization (column 1), we randomly select 15 categories from different super classes in ImageNet-1k following WordNet.  For the fine-grained feature visualization (column 2), we randomly select 11 fine-grain categories under 3 different super classes. For the failure case visualization, we select the categories which have the low in-domain accuracy.}
    \label{fig:feature_space}
% \vspace{-0.4cm}
\end{figure*}

\end{document}